\pdfoutput=1
\documentclass[journal]{IEEEtran}

\usepackage{amsmath,amssymb}
\usepackage{graphicx}
\usepackage{booktabs}
\usepackage{algorithm}
\usepackage{algorithmic}
\usepackage{xcolor}
\usepackage{tikz}
\usetikzlibrary{arrows.meta,positioning,calc,decorations.pathmorphing}
\usepackage[hidelinks]{hyperref}

\definecolor{cfblue}{RGB}{31,90,158}
\definecolor{cfteal}{RGB}{0,140,150}
\definecolor{cfgray}{RGB}{90,98,110}
\definecolor{cforange}{RGB}{205,110,40}
\definecolor{cfred}{RGB}{180,50,55}
\definecolor{cfbg}{RGB}{238,242,247}

\graphicspath{{./figures/}}

\begin{document}

\title{CFRNet: Cycle-Consistent Fixed-Point Training for Real-Time Blind Face Restoration on Consumer Embedded NPUs}

\author{Fuchen~Li,~Xinyang~Wang,~Yahui~Zhang,~Yuhan~Chen,~Jiahong~Guo,~Zhuohan~Qin,~and~Wenbo~Ma% 
\thanks{F. Li and X. Wang contributed equally to this work. \textit{(Corresponding author: Fuchen Li, (e-mail: fuchen.li@ufl.edu).)}}%
\thanks{F. Li, X. Wang, and J. Guo are with the University of Florida, Gainesville, FL, USA.}%
\thanks{Y. Zhang is with the University of Southampton, Southampton, UK.}%
\thanks{Y. Chen is with Chongqing University, Chongqing, China.}%
\thanks{Z. Qin is with Qingdao University, Qingdao, China.}%
\thanks{W. Ma is with Intel Asia-Pacific Research \& Development Ltd, Shanghai, China.}}

% \markboth{IEEE Transactions on Consumer Electronics}%
% {Li \MakeLowercase{\textit{et al.}}: CFRNet: Cycle-Consistent Fixed-Point Training for Blind Face Restoration on Consumer Embedded NPUs}

\maketitle

\begin{abstract}
Blind face restoration on consumer devices has to balance image quality against speed and memory.
Strong methods such as GFPGAN and CodeFormer give good perceptual quality, but they rely on large pretrained generative priors and on operators such as attention, codebook lookup, and style modulation that are hard to compile and quantize on the small neural processing units (NPUs) used in consumer hardware.
Small convolutional restorers run fast enough, but they tend to over-smooth and to leave artifacts around the eyes, nose, and mouth.
We present CFRNet, a 2.0\,M-parameter ResNet-style restorer for on-device use at $256\times256$, the common face-crop size on consumer NPUs.
The main idea is Cycle-Consistent Fixed-Point Training (CCFP). Instead of training the network for one pass and then running it several times by hand, we train it to act as a fixed-point operator, so that applying it again to a restored face does not change the face.
CCFP uses three training losses, namely progressive multi-cycle supervision, an idempotence loss, and a re-degradation cycle loss, and it adds no cost at inference.
To compare fairly under our deployment limits, we retrain all baselines from scratch at the same $256\times256$ resolution.
On a 300-image test set, CFRNet reaches the best perceptual score (LPIPS 0.250 at three cycles, which is 31\% lower than one cycle) and also the best PSNR and SSIM at two cycles.
It runs in about 23\,ms per cycle in INT8 on a HiSilicon Hi3402 NPU, while the same baselines cannot be compiled to that chip.
The cycle count $k$ acts as a simple quality knob that needs no retraining: PSNR is best at $k\!=\!2$ and LPIPS keeps improving up to $k\!=\!3$.
We further show that the same idea works with a plain CNN that is even easier to deploy, and we run the model in real time on an in-car driver-monitoring board.
\end{abstract}


% ============================================================
\section{Introduction}
% ============================================================
\IEEEPARstart{B}{lind} face restoration takes a degraded face image, which may have blur, downsampling, noise, and compression artifacts, and recovers a clean and identity-consistent face.
The task matters for many consumer products: photo and video apps on phones, video calls, smart cameras and doorbells, and in-car driver-monitoring systems~\cite{reddy2017drowsiness,yu2019drowsiness}.
In these products the input is often captured in poor conditions, and the model has to run on a small NPU with a tight latency and memory budget.
Recent prior-guided methods such as GFPGAN~\cite{wang2021gfpgan}, GPEN~\cite{yang2021gpen}, RestoreFormer~\cite{wang2022restoreformer}, and CodeFormer~\cite{zhou2022codeformer} give very good perceptual quality.
They do this with pretrained StyleGAN2 generators, transformer modules, or learned codebooks. The cost is size and complexity.
Counting the frozen priors, these systems often have 70--85\,M parameters, and they use operators such as style modulation, multi-head attention, and codebook lookup.
These operators are hard to compile and to quantize to INT8 on the NPUs found in consumer devices.
Small CNN restorers~\cite{sandler2018mobilenetv2,liu2020rfdn} fit the latency budget, but they have two common problems: they over-smooth high-frequency texture, and they leave artifacts around the eyes, nose, and mouth.
One way to make a small network better without making it larger is to run it more than once at test time.
Earlier work on iterative super-resolution~\cite{haris2018dbpn,ma2020dic} and recurrent restoration~\cite{lee2019recurrent} ran a single network several times.
But these methods train the network for a single forward pass and then reuse it in a heuristic way.
This leaves a basic question open: why should running a single-pass network again make the result better instead of worse?
What stops the second pass from harming a face that is already restored?
If $G$ is trained only to map a degraded image $x$ to a clean image $y$, then feeding it $y'\approx y$ is out of distribution.
The network was never asked to keep a good image good, so further passes may over-sharpen, add detail that is not there, or oscillate.
We call this the training-inference gap.

This paper closes that gap.
We present CFRNet, a small restorer for consumer NPUs, and Cycle-Consistent Fixed-Point Training (CCFP).
Rather than train $G$ for one mapping and reuse it by hand, CCFP trains $G$ to be a fixed-point operator on the set of natural faces.
Three losses work together:
\begin{itemize}
\item Progressive supervision. We supervise the output of every cycle $G^{(i)}(x)$, $i=1,2,3$, against the ground truth, with larger weight on later cycles.
This teaches $G$ a coarse-to-fine refinement instead of a one-shot mapping.
\item Idempotence.
A loss $\lVert G(G^{(k)}(x))-G^{(k)}(x)\rVert_1$ pushes the last output toward a fixed point of $G$, so one more pass does not change it.
\item Re-degradation cycle. If we degrade a restored face again with the degradation model $D$ and restore it, we should get the same restored face: $G(D(G^{(k)}(x)))\approx G^{(k)}(x)$.
This pushes $G$ to learn the inverse of the degradation process, not to memorize training pairs.
\end{itemize}

CCFP only changes training.
The deployed network is the same small model, run $k=3$ times.
The difference is that its iterative behavior is now designed, not accidental.
Our target chip is the HiSilicon Hi3402 NPU, a part used in consumer and automotive vision products.
It compiles INT8 graphs of standard CNN operators and runs at a $256\times256$ face crop.
The official GFPGAN, GPEN, and CodeFormer releases run at $512\times512$ and use operators that the Hi3402 toolchain does not support or that lose too much quality in INT8.
To make a fair comparison under these limits, we reimplement each baseline as a deploy-compatible ``Lite'' model with no frozen prior and no transformer or codebook modules, and we train all of them from scratch on the same data, the same degradation, the same $256\times256$ resolution, and the same schedule as CFRNet.
We do not claim that CFRNet beats the full official models at $512\times512$.
We show that, under limits that a consumer NPU can actually meet, CFRNet gives the best perceptual quality among comparable from-scratch CNN baselines.
We also observe that, on the same inputs, CFRNet at $k\!=\!3$ reaches about the same perceptual quality as the full official GFPGAN, while using a small fraction of its parameters and compute (Section~\ref{subsec:quant_main}).
Our contributions are as follows.
\begin{enumerate}
\item CCFP, a training method that links cyclical inference to a fixed-point objective on the natural-face set.
It combines progressive multi-cycle supervision, idempotence, and a re-degradation cycle in one framework.
\item A small ResNet-style generator with 2.0\,M parameters that uses only standard CNN operators, runs in INT8, and takes about 23\,ms per cycle on the Hi3402 NPU.
\item A nose region added to the usual GFPGAN-style component supervision, which removes a common mid-face artifact of small CNN restorers.
\item A perception-distortion trade-off across cycles: PSNR is best at $k\!=\!2$, while LPIPS keeps improving up to $k\!=\!3$.
The cycle count is a quality knob that needs no retraining.
\item Evidence that the same iterative idea transfers to a plain, easy-to-deploy CNN that reaches $k\!=\!3$ quality on hard real inputs, plus a real-time in-car deployment on the embedded board.
\end{enumerate}

% ============================================================
\section{Related Work}
% ============================================================

\subsection{GAN-based and prior-guided face restoration}
Early CNN restorers improved fidelity~\cite{dong2016srcnn,lim2017edsr,zhang2018rcan,liang2021swinir} but gave over-smooth outputs under strong degradation.
SRGAN and ESRGAN showed that adversarial training helps perceptual sharpness~\cite{goodfellow2014gan,ledig2017photo,wang2018esrgan}, and face-specific methods used geometric or identity priors~\cite{bulat2018superfan,chen2018fsrnet}.
Prior-guided methods use strong generators: GFPGAN injects features from a pretrained StyleGAN2~\cite{wang2021gfpgan}, GPEN embeds a GAN prior in a U-shaped decoder~\cite{yang2021gpen}, and RestoreFormer~\cite{wang2022restoreformer} and CodeFormer~\cite{zhou2022codeformer} use transformers and discrete codebooks~\cite{gu2022vqfr}.
These give strong perceptual quality, but the full systems are large and their main modules do not map well to consumer NPU compilation or INT8.
\subsection{Component-aware supervision}
Local supervision on face parts appears in landmark- and parsing-guided methods~\cite{chen2018fsrnet} and in component-dictionary methods~\cite{li2020dfdnet}.
GFPGAN made local discriminators on the left eye, right eye, and mouth popular~\cite{wang2021gfpgan}.
We add a nose region, which we find reduces mid-face artifacts in small generators.
\subsection{Iterative and recurrent restoration}
Iterative and recurrent designs are common in super-resolution.
Back-projection networks~\cite{haris2018dbpn} alternate up- and down-projection blocks, DIC~\cite{ma2020dic} refines face detail in a recurrent way, and some methods reapply a small generator several times~\cite{lee2019recurrent}.
Diffusion-based restorers~\cite{saharia2022sr3,yue2024difface} and direct-iteration models~\cite{delbracio2023indi} also treat restoration as repeated steps.
The closest work to ours~\cite{lee2019recurrent} trains a small network with a few recurrent passes, but it supervises only the final output and does not treat $G$ as a fixed-point operator.
CFRNet is different in two ways: we supervise every cycle with increasing weights, and we add an idempotence loss and a re-degradation cycle loss that tie the procedure to a fixed point.
\subsection{Fixed-point and cycle objectives}
Deep equilibrium models~\cite{bai2019deq} and consistency models~\cite{song2023consistency} use fixed-point ideas, but they usually need an iterative solver in the forward pass and are heavy.
Cycle consistency is widely used in unpaired translation~\cite{zhu2017cyclegan}, and re-degradation cycles appear in recent reference-based face restoration~\cite{refstar2025} and unsupervised diffusion-prior restoration~\cite{kuai2024unsupervised}.
To our knowledge, no prior face-restoration work combines a small feed-forward CNN, weight-shared multi-cycle inference, and joint fixed-point and re-degradation training.
\subsection{The perception-distortion trade-off}
Blau and Michaeli~\cite{blau2018perception} show that pixel distortion (such as PSNR) and perceptual quality (such as LPIPS or FID) cannot both be minimized at once.
Later face-restoration work puts perceptual metrics first. Our cycle analysis (Section~\ref{subsec:cycle_ablation}) shows this trade-off inside one trained CFRNet: the cycle that is best for PSNR is not the cycle that is best for LPIPS.
\subsection{Efficient networks for on-device restoration}
On-device restoration usually uses small backbones and depthwise-separable operators~\cite{sandler2018mobilenetv2,howard2019mobilenetv3,tan2019efficientnet,liu2020rfdn}.
CFRNet uses a simple ResNet-style encoder-decoder with standard convolutions, ReLU, and nearest-neighbor upsampling, which are all supported by embedded NPU toolchains and quantize to INT8~\cite{jacob2018quantization} without much loss.
% ============================================================
\section{Method}
% ============================================================
\label{sec:method}

\subsection{Overview}
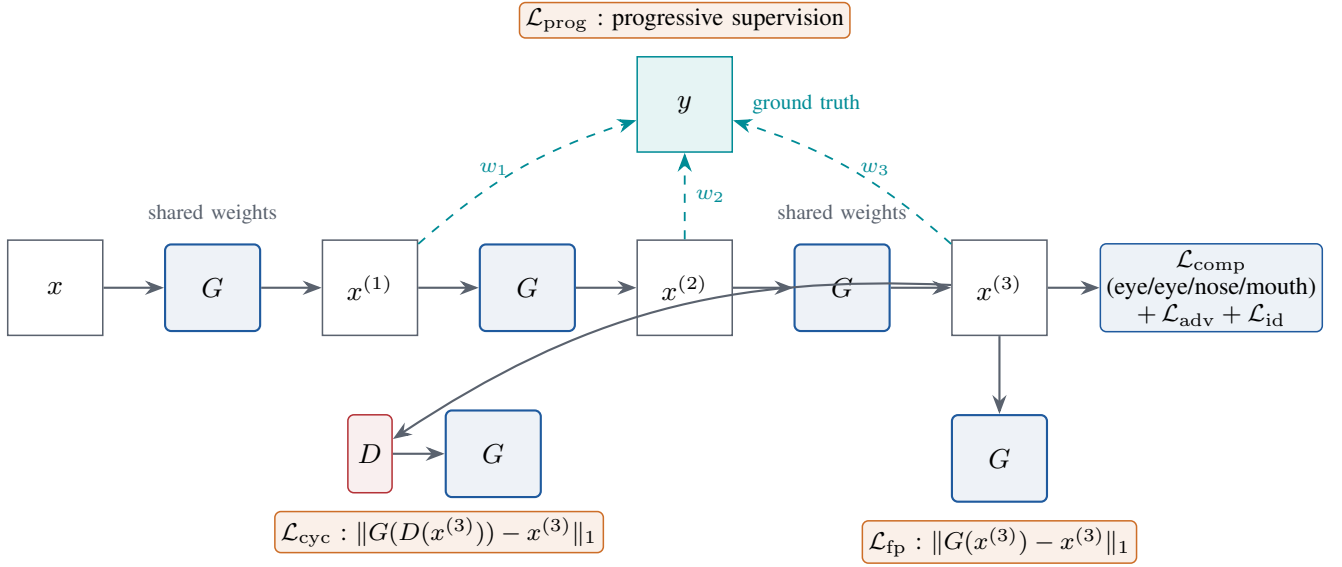
\begin{figure*}[t]
\centering
\resizebox{0.96\linewidth}{!}{%
\begin{tikzpicture}[
  font=\small,
  >={Stealth[length=2.2mm]},
  gen/.style={draw=cfblue,fill=cfbg,line width=0.7pt,rounded corners=2pt,minimum width=11mm,minimum height=10mm,align=center},
  img/.style={draw=cfgray,fill=white,line width=0.5pt,minimum width=11mm,minimum height=11mm,align=center},
  lossn/.style={draw=cforange,fill=cforange!10,line width=0.5pt,rounded corners=2pt,align=center,font=\footnotesize,inner sep=2pt},
  flow/.style={->,line width=0.7pt,cfgray},
]
\node[img] (x0) {$x$};
\node[gen,right=7mm of x0] (g1) {$G$};
\node[img,right=7mm of g1] (x1) {$x^{(1)}$};
\node[gen,right=7mm of x1] (g2) {$G$};
\node[img,right=7mm of g2] (x2) {$x^{(2)}$};
\node[gen,right=7mm of x2] (g3) {$G$};
\node[img,right=7mm of g3] (x3) {$x^{(3)}$};
\draw[flow] (x0)--(g1); \draw[flow] (g1)--(x1);
\draw[flow] (x1)--(g2); \draw[flow] (g2)--(x2);
\draw[flow] (x2)--(g3);
\draw[flow] (g3)--(x3);
\node[above=1mm of g1,font=\scriptsize,cfgray] {shared weights};
\node[above=1mm of g3,font=\scriptsize,cfgray] {shared weights};
\node[img,above=10mm of x2,fill=cfteal!10,draw=cfteal] (y) {$y$};
\node[right=1mm of y,font=\scriptsize,cfteal] {ground truth};
\draw[->,cfteal,dashed,line width=0.6pt] (x1) to[bend left=12] node[midway,left,font=\scriptsize]{$w_1$} (y);
\draw[->,cfteal,dashed,line width=0.6pt] (x2) -- node[midway,right,font=\scriptsize]{$w_2$} (y);
\draw[->,cfteal,dashed,line width=0.6pt] (x3) to[bend right=12] node[midway,right,font=\scriptsize]{$w_3$} (y);
\node[lossn,above=2mm of y] (lprog) {$\mathcal{L}_{\mathrm{prog}}$ : progressive supervision};
\node[gen,below=9mm of x3] (gfp) {$G$};
\draw[flow] (x3) -- (gfp);
\node[lossn,below=2mm of gfp] (lfp) {$\mathcal{L}_{\mathrm{fp}}$ : $\|G(x^{(3)})-x^{(3)}\|_1$};
\node[draw=cfred,fill=cfred!8,rounded corners=2pt,line width=0.5pt,below=9mm of x1,minimum height=9mm] (dd) {$D$};
\node[gen,right=6mm of dd] (gc) {$G$};
\draw[flow] (x3) to[bend right=18] (dd);
\draw[flow] (dd) -- (gc);
\node[lossn,below=2mm of dd,xshift=8mm] (lcyc) {$\mathcal{L}_{\mathrm{cyc}}$ : $\|G(D(x^{(3)}))-x^{(3)}\|_1$};
\node[lossn,right=6mm of x3,fill=cfblue!8,draw=cfblue] (lcomp) {$\mathcal{L}_{\mathrm{comp}}$\\(eye/eye/nose/mouth)\\$+\,\mathcal{L}_{\mathrm{adv}}+\mathcal{L}_{\mathrm{id}}$};
\draw[flow] (x3) -- (lcomp);
\end{tikzpicture}}
\caption{CFRNet and CCFP training.
The same 2.0\,M-parameter generator $G$ is run $k=3$ times with shared weights.
Progressive supervision compares each cycle output $x^{(i)}$ to the ground truth $y$ with increasing weights $w_1<w_2<w_3$.
The idempotence loss pushes the last output toward a fixed point of $G$.
The re-degradation cycle loss degrades the restored image with $D$ and restores it again, and asks the result to match the original restoration.
Local discriminators on the eyes, nose, and mouth, together with adversarial and identity losses, supervise the final cycle.
At inference only the top row runs; no extra parameters or operators are added.}
\label{fig:overview}
\end{figure*}

Given a degraded face $x\in\mathbb{R}^{H\times W\times3}$ and its clean version $y$, we want a small generator $G$ such that running $G$ a few times gives a good restoration:
\begin{equation}
\hat{y}=G^{(k)}(x):=\underbrace{G\circ G\circ\cdots\circ G}_{k\text{ times}}(x).
\end{equation}
Fig.~\ref{fig:overview} shows CFRNet. It has three parts: a small generator (Section~\ref{subsec:generator}), component-aware local supervision (Section~\ref{subsec:component}), and CCFP training (Section~\ref{subsec:ccfp}).
% ---------- TikZ overview figure (double column) ----------

\subsection{Small Generator}
\label{subsec:generator}
The generator is a ResNet-style encoder-decoder (Table~\ref{tab:gen_arch}) with base width $C=32$, two stride-2 downsampling stages, $N=6$ residual blocks at the bottleneck, and two upsampling stages that use nearest-neighbor upsampling with $3\times3$ convolutions.
Skip connections join matching encoder and decoder stages. The output head predicts a residual that is added to the input before a $\tanh$: $\hat{y}=\tanh(x+\mathrm{head}(\cdot))$.
The residual design is important for stable training. With small initial head weights the network starts near the identity, which avoids the constant-output collapse we saw with direct image regression.
The backbone uses only convolution, ReLU, residual addition, and nearest-neighbor upsampling.
All of these are supported by embedded NPU toolchains and quantize to INT8 cleanly.

\begin{table}[t]
\centering
\caption{Generator configuration ($H=W=256$).
Total: \textbf{2.00\,M parameters}, \textbf{8.7\,G MACs per cycle}.}
\label{tab:gen_arch}
\begin{tabular}{lccc}
\toprule
Stage & Operator & Output size & Channels \\
\midrule
Input  & --                                & $256\times256$ & 3 \\
Stem   & $3\times3$ conv                    & $256\times256$ & 32 \\
Down-1 & $3\times3$ conv, s\,=\,2           & $128\times128$ & 
64 \\
Down-2 & $3\times3$ conv, s\,=\,2           & $64\times64$   & 128 \\
Body   & ResBlock $\times\,6$              & $64\times64$   & 128 \\
Up-1   & conv $+$ nn-up                     & $128\times128$ & 64 \\
Up-2   & conv $+$ nn-up                
     & $256\times256$ & 32 \\
Head   & $3\times3$ conv $+$ residual add $+$ \texttt{tanh} & $256\times256$ & 3 \\
\bottomrule
\end{tabular}
\end{table}

\subsection{Component-Aware Supervision}
\label{subsec:component}
Following GFPGAN~\cite{wang2021gfpgan}, we put local discriminators on aligned face-component patches.
We add a nose region to the GFPGAN set:
\begin{equation}
\mathcal{R}=\{\mathrm{le},\mathrm{re},\mathrm{n},\mathrm{m}\}.
\end{equation}
For each region $r$, ROI Align crops the restored patch $\hat{y}_r$ and the ground-truth patch $y_r$, and a small local discriminator $D_r$ is trained with them.
The component loss is a local hinge adversarial term plus a feature-style (Gram) term:
\begin{equation}
\begin{aligned}
\mathcal{L}_{\mathrm{comp}}=\sum_{r\in\mathcal{R}}\Big[
& -\lambda_{\mathrm{loc}}\,\mathbb{E}_{\hat{y}_r}\big[D_r(\hat{y}_r)\big] \\
& +\lambda_{\mathrm{fs}}\,\|\mathrm{Gram}(\psi_r(\hat{y}_r))-\mathrm{Gram}(\psi_r(y_r))\|_1\Big],
\end{aligned}
\end{equation}
where $\psi_r$ are features from $D_r$.
The discriminators use spectral normalization. The nose region reduces ringing around the nostrils and the nasal bridge, which is a common failure of small CNN restorers.
\subsection{Cycle-Consistent Fixed-Point Training}
\label{subsec:ccfp}

\subsubsection{Motivation}
\begin{figure}[t]
\centering
\resizebox{0.98\linewidth}{!}{%
\begin{tikzpicture}[font=\small,>={Stealth[length=2.2mm]}]
  \def\W{7.8} \def\H{4.6}
  \draw[->,line width=0.8pt] (0,0) -- (\W,0);
\draw[->,line width=0.8pt] (0,0) -- (0,\H);
  \node[below=4mm] at (\W/2,0) {Inference cycle $k$};
  \node[rotate=90] at (-1.15,\H/2) {Distance to clean image $\lVert x^{(k)}-y\rVert$};
\foreach \x/\lab in {1.5/1,3.0/2,4.5/3,6.0/4}{
    \draw (\x,0.06) -- (\x,-0.06);
    \node[below,font=\footnotesize] at (\x,-0.06) {\lab};
}
  \node[below left,font=\footnotesize] at (0,0) {0};
  \node[anchor=east,font=\footnotesize] at (-0.18,0.6) {low};
  \node[anchor=east,font=\footnotesize] at (-0.18,3.6) {high};
  \draw (0.06,0.6) -- (-0.06,0.6);
\draw (0.06,3.6) -- (-0.06,3.6);
  \draw[dashed,cfteal,line width=0.7pt] (0,0.6) -- (\W-0.2,0.6);
  \node[cfteal,font=\footnotesize,anchor=east] at (\W-0.1,0.9) {natural-face manifold (fixed point)};
\draw[cfred,line width=1.1pt]
    (0,3.6) .. controls (0.8,2.4) and (1.2,1.9) .. (1.5,1.7)
            .. controls (2.2,1.4) and (2.6,1.35) .. (3.0,1.45)
            .. controls (3.8,1.7) and (4.2,2.0) .. (4.5,2.35)
            .. controls (5.2,2.9) and (5.6,3.1) .. (6.0,3.25);
\node[cfred,font=\footnotesize,anchor=west] at (4.55,3.05) {heuristic: drifts, over-sharpens};
  \fill[cfred] (1.5,1.7) circle (1.3pt) (3.0,1.45) circle (1.3pt) (4.5,2.35) circle (1.3pt) (6.0,3.25) circle (1.3pt);
\draw[cfblue,line width=1.1pt]
    (0,3.6) .. controls (0.8,2.3) and (1.2,1.8) .. (1.5,1.55)
            .. controls (2.2,1.1) and (2.6,0.85) .. (3.0,0.72)
            .. controls (3.8,0.63) and (4.2,0.61) .. (4.5,0.6)
            .. controls (5.2,0.6) and (5.6,0.6) .. (6.0,0.6);
\node[cfblue,font=\footnotesize,anchor=west] at (2.45,1.65) {CCFP: converges and stays};
  \fill[cfblue] (1.5,1.55) circle (1.3pt) (3.0,0.72) circle (1.3pt) (4.5,0.6) circle (1.3pt) (6.0,0.6) circle (1.3pt);
\end{tikzpicture}}
\caption{Distance to the clean image as a function of the cycle count.
A single-pass network reused by hand (red) improves for one or two passes and then drifts back up, because its own outputs are out of distribution.
CCFP (blue) keeps the distance going down and then flat, which is the fixed-point behavior.
The matching numbers are in Table~\ref{tab:fixedpoint}.}
\label{fig:dynamics}
\end{figure}
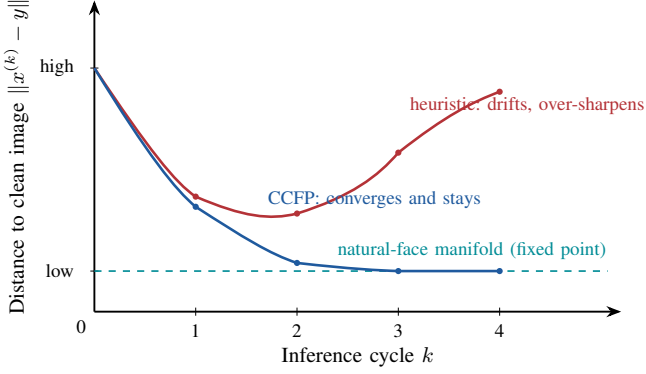

A single-pass loss $\mathcal{L}(G(x),y)$ learns the map from a degraded image to a clean one.
It says nothing about how $G$ behaves on its own outputs.
So when we run $G$ again at test time, two things can go wrong: the second pass can over-sharpen or add fake texture because $G(x^{(1)})$ is out of distribution, or the iterations can oscillate and never settle.
CCFP treats $G$ as a discrete dynamical system on image space and trains it to move toward the set of natural faces $\mathcal{M}_y$ and to stop there (Fig.~\ref{fig:dynamics}).
The clean image $y$ is the target fixed point, and we want each trajectory $\{x^{(0)}=x,x^{(1)},x^{(2)},\dots\}$ to converge near $\mathcal{M}_y$ and stay.
% ---------- TikZ dynamics figure (single column, with axes) ----------

\subsubsection{The three CCFP losses}
Let $x^{(0)}=x$ and $x^{(i)}=G(x^{(i-1)})$ for $i=1,\dots,k$. We use $k=3$.
\paragraph{Progressive multi-cycle supervision}
Each cycle output is supervised against $y$ with increasing weight:
\begin{equation}
\mathcal{L}_{\mathrm{prog}}=\sum_{i=1}^{k}w_i\big[\|x^{(i)}-y\|_1+\lambda_{\mathrm{per}}\|\phi(x^{(i)})-\phi(y)\|_1\big],
\label{eq:prog}
\end{equation}
where $\phi$ is a pretrained VGG-19 feature extractor~\cite{johnson2016perceptual} and $w_1=0.3$, $w_2=0.5$, $w_3=1.0$.
The later cycle gets more weight because it is the deployment output, and the earlier cycles are still supervised so that each pass gives a usable image.
This loss replaces the usual single-pass reconstruction loss.

\paragraph{Idempotence loss}
The last cycle output should be a fixed point of $G$:
\begin{equation}
\mathcal{L}_{\mathrm{fp}}=\|G(x^{(k)})-x^{(k)}\|_1.
\label{eq:fp}
\end{equation}
This costs one extra forward pass at training time. It says that one more pass should not change the output, which prevents over-sharpening and gives a natural stopping point.
In other words, it makes $\mathcal{M}_y$ a stable attractor of $G$.
\paragraph{Re-degradation cycle loss}
Let $D$ be a random degradation simulator that follows the training degradation.
Degrading a restored image and restoring it again should give the same restoration:
\begin{equation}
\mathcal{L}_{\mathrm{cyc}}=\|G(D(x^{(k)}))-x^{(k)}\|_1.
\label{eq:cyc}
\end{equation}
Where $\mathcal{L}_{\mathrm{prog}}$ teaches $G$ to map $x$ to $y$, this loss teaches $G$ to undo the degradation applied to a clean image.
Because $D$ is random, this term also works as a regularizer and helps on harder degradations.
\subsubsection{Full objective}
The full training loss is
\begin{equation}
\mathcal{L}_{\mathrm{total}}=\mathcal{L}_{\mathrm{prog}}+\lambda_{\mathrm{fp}}\mathcal{L}_{\mathrm{fp}}+\lambda_{\mathrm{cyc}}\mathcal{L}_{\mathrm{cyc}}+\lambda_{\mathrm{adv}}\mathcal{L}_{\mathrm{adv}}+\mathcal{L}_{\mathrm{comp}}+\lambda_{\mathrm{id}}\mathcal{L}_{\mathrm{id}},
\label{eq:total}
\end{equation}
where $\mathcal{L}_{\mathrm{adv}}$ is a global hinge GAN loss on $x^{(k)}$ only, $\mathcal{L}_{\mathrm{id}}$ is an ArcFace identity loss~\cite{deng2019arcface}, and $\mathcal{L}_{\mathrm{comp}}$ is the component loss above.
We use $\lambda_{\mathrm{per}}=1$, $\lambda_{\mathrm{fp}}=0.5$, $\lambda_{\mathrm{cyc}}=0.5$, $\lambda_{\mathrm{adv}}=0.1$, $\lambda_{\mathrm{loc}}=1$, $\lambda_{\mathrm{fs}}=1$, $\lambda_{\mathrm{id}}=10$.

\subsubsection{Training stability}
We raise $\lambda_{\mathrm{fp}}$ from $0$ to $0.5$ over the first 100K iterations.
If it is large from the start, the idempotence loss can collapse $G$ to the identity, which trivially gives $G(G(x))=G(x)$.
We also use a 5{,}000-iteration GAN warm-up, where only the reconstruction terms drive training, followed by a 5{,}000-iteration ramp on the adversarial and component-discriminator losses.
The discriminators use spectral normalization and a hinge loss, and we clip the gradient norm at 1.0.
These choices were needed: without them, early runs collapsed to a constant image.
\begin{algorithm}[t]
\caption{One CCFP training step}
\label{alg:ccfp}
\begin{algorithmic}[1]
\REQUIRE clean image $y$, degradation $D$, generator $G$, cycles $k=3$
\STATE $x\leftarrow D(y)$ \hfill // sample a degraded input
\STATE $x^{(0)}\leftarrow x$
\FOR{$i=1$ to $k$}
  \STATE $x^{(i)}\leftarrow G(x^{(i-1)})$
\ENDFOR
\STATE $x^{(\mathrm{fp})}\leftarrow G(x^{(k)})$ \hfill // for the idempotence loss
\STATE $x^{(\mathrm{cyc})}\leftarrow G(D(x^{(k)}))$ \hfill // for the cycle loss
\STATE compute $\mathcal{L}_{\mathrm{prog}}$ from $\{x^{(i)}\}$ and $y$
\STATE $\mathcal{L}_{\mathrm{fp}}\leftarrow\|x^{(\mathrm{fp})}-x^{(k)}\|_1$
\STATE $\mathcal{L}_{\mathrm{cyc}}\leftarrow\|x^{(\mathrm{cyc})}-x^{(k)}\|_1$
\STATE compute $\mathcal{L}_{\mathrm{adv}},\mathcal{L}_{\mathrm{comp}},\mathcal{L}_{\mathrm{id}}$ from $x^{(k)}$ and $y$
\STATE backprop $\mathcal{L}_{\mathrm{total}}$ and update $G$
\end{algorithmic}
\end{algorithm}

\subsubsection{Inference}
At test time we run $G$ for $k$ cycles with shared weights, $\hat{y}=G^{(k)}(x)$.
We do not use the idempotence pass or the degradation simulator at test time, since both are training-only.
Latency grows linearly with $k$. We study the effect of $k$ in Section~\ref{subsec:cycle_ablation} and use $k=3$ by default.
% ============================================================
\section{Experiments}
% ============================================================
\label{sec:exp}

\subsection{Setup}
\label{subsec:protocol}

\paragraph{Training data}
We train on FFHQ~\cite{karras2019stylegan}, which has 70K aligned high-quality faces. Images are resized to $256\times256$.
We use a fixed 90/5/5\% split: 63K for training, 3.5K for validation, and 3.5K held out.
\paragraph{Degradation $D$}
Following GFPGAN~\cite{wang2021gfpgan,wang2021realesrgan}, $D$ applies four steps with random parameters: Gaussian blur with $\sigma\in[1,15]$;
bicubic downsampling by a random scale $r\in[1,12]$ and bicubic upsampling back to $256\times256$;
additive Gaussian noise with $\sigma_n\in[0,20]$ on the $[0,255]$ scale; and JPEG compression with quality $q\in[30,90]$.
The same $D$ is used for the re-degradation cycle loss.
For evaluation we fix the random seeds per image, so every method sees the same degraded inputs.
\paragraph{Evaluation set}
We use a fixed test set of 300 FFHQ images that are held out from training.
The same per-image degradation is applied to all methods, so the inputs are identical across the methods we compare.
We report PSNR, SSIM, and LPIPS averaged over the 300 images.
\paragraph{Baselines}
\label{par:baselines}
For a fair comparison under our deployment limits ($256\times256$ crops, INT8 on the Hi3402, no pretrained prior), we reimplement each method as a deploy-compatible model trained from scratch on the same data:
\begin{itemize}
\item \textbf{GFPGAN-Lite} ($\sim$17\,M): a U-Net restorer that follows GFPGAN's degradation-removal module~\cite{wang2021gfpgan} but drops the frozen StyleGAN2 prior and the SFT modules, which are not NPU-compilable.
Trained with the same losses as GFPGAN (L1, perceptual, adversarial, identity, and eye/mouth component discriminators).
\item \textbf{GPEN-Lite} ($\sim$15\,M): a style-modulated U-Net inspired by GPEN~\cite{yang2021gpen}, without GPEN's full StyleGAN2 decoder.
Trained with reconstruction, adversarial, and identity losses.
\item \textbf{CodeFormer-Lite} ($\sim$10.5\,M): a VQ-VAE-style restorer with a 1024-entry codebook, trained end to end with reconstruction, adversarial, and commitment losses.
It drops the two-stage codebook pretraining and the transformer code prediction of the official CodeFormer~\cite{zhou2022codeformer}, which do not compile on the NPU.
\end{itemize}
This setup separates the effect of each design's architecture from the effect of large pretrained priors that cannot run on the target NPU.
We do not compare against the full $512\times512$ official models, except for the qualitative parity note in Section~\ref{subsec:quant_main}.
\paragraph{Metrics}
We report PSNR and SSIM for distortion and LPIPS~\cite{zhang2018lpips} for perceptual quality.
For blind face restoration the perceptual metric is the main one~\cite{blau2018perception};
PSNR and SSIM are given for completeness and to show the perception-distortion trade-off (Section~\ref{subsec:cycle_ablation}).

\paragraph{Implementation}
Generator: $C=32$, $N=6$ (Table~\ref{tab:gen_arch}).
Global discriminator: a PatchGAN~\cite{isola2017pix2pix} with spectral normalization~\cite{miyato2018spectral}. Component discriminators: 4-layer SN-Conv-LeakyReLU networks on ROI crops ($40\times40$ for eyes and nose, $80\times50$ for the mouth).
We use Adam ($\text{lr}=2\times10^{-4}$, $\beta_1=0.5$, $\beta_2=0.999$) and batch size 16. CFRNet is trained for 32 epochs at the base learning rate and then fine-tuned with the learning rate scaled by $0.25$.
All baselines use the same hardware (one NVIDIA L40S) and the same number of iterations.
\paragraph{Latency measurement}
We report latency on two platforms. The server number is measured on one NVIDIA Tesla V100 in FP32 with batch size 1, using CUDA events and 50 timed runs after 10 warm-up runs.
The board number is measured on the HiSilicon Hi3402 NPU in INT8.
The two numbers come from different hardware and precision, so they are not directly comparable;
we give both to be clear about what runs where.

\subsection{Main Comparison}
\label{subsec:quant_main}

Table~\ref{tab:quant_main} reports the comparison under the shared protocol.
The degraded input has PSNR 22.45 and LPIPS 0.712 as a reference floor.

\begin{table*}[t]
\centering
\caption{Comparison on the 300-image FFHQ-256 test set.
All baselines are reimplemented as deploy-compatible ``Lite'' models and trained from scratch at $256\times256$ with the same degradation, schedule, and hardware as CFRNet.
``Board'' latency is on the HiSilicon Hi3402 NPU in INT8; \emph{infeasible} means the model cannot be compiled to the NPU.
``Server'' latency is on one V100 in FP32. Best in \textbf{bold}, second \underline{underlined}. CFRNet variants share the same 2.0\,M parameters;
only the cycle count differs.}
\label{tab:quant_main}
\begin{tabular}{lccccccc}
\toprule
Method & Params $\downarrow$ & MACs/pass $\downarrow$ & PSNR $\uparrow$ & SSIM $\uparrow$ & LPIPS $\downarrow$ & Server FP32 (ms) & Board INT8 (ms) \\
\midrule
Input (degraded)            & --            & --              & 22.45 & 0.621 & 0.712 & -- & -- \\
\midrule
GFPGAN-Lite~\cite{wang2021gfpgan}   & $\sim$17.0\,M & $\sim$23\,G   & 23.49 & 0.668 & 0.353 & -- & infeasible \\
GPEN-Lite~\cite{yang2021gpen}  
     & $\sim$15.0\,M & $\sim$19\,G   & 23.18 & 0.659 & 0.378 & -- & infeasible \\
CodeFormer-Lite~\cite{zhou2022codeformer} & $\sim$10.5\,M & $\sim$11\,G & 22.71 & 0.642 & 0.382 & -- & infeasible \\
\midrule
CFRNet ($k\!=\!1$)         & \textbf{2.0\,M} & \textbf{8.7\,G}  & 23.54 & 0.673 & 0.360 & \textbf{2.7} & \textbf{23} \\
CFRNet ($k\!=\!2$)         & \textbf{2.0\,M} & 17.4\,G          & \textbf{23.82} & \textbf{0.685} & 0.300 & 4.9 & 46 \\
CFRNet ($k\!=\!3$, default) & \textbf{2.0\,M} & 
26.1\,G          & 23.10 & 0.662 & \textbf{0.250} & 7.3 & 69 \\
CFRNet ($k\!=\!4$)         & \textbf{2.0\,M} & 34.8\,G          & 22.98 & 0.668 & \underline{0.262} & 9.6 & 92 \\
\bottomrule
\end{tabular}
\end{table*}

There are three main points.
First, CFRNet gives the best perceptual score by a clear margin.
Its LPIPS at $k\!=\!3$ (0.250) is 29\% lower than GFPGAN-Lite (0.353), 34\% lower than GPEN-Lite, and 35\% lower than CodeFormer-Lite, with 5 to 8 times fewer parameters.
Because LPIPS is the main metric for this task~\cite{blau2018perception,wang2021gfpgan,zhou2022codeformer}, this is the headline result.
CFRNet at $k\!=\!2$ also gives the best PSNR (23.82) and the best SSIM (0.685), so it is not trading away pixel fidelity to get there.
Second, CFRNet is the only model that runs on the NPU.
The Lite baselines still keep operators that the Hi3402 INT8 toolchain cannot compile, even after we removed the heavy prior modules.
CFRNet uses only standard CNN operators and runs end to end at 69\,ms for $k\!=\!3$.
Third, PSNR and SSIM behave differently from LPIPS across cycles.
CFRNet's PSNR is highest at $k\!=\!2$ (23.82) and lower at $k\!=\!3$ (23.10) and $k\!=\!4$ (22.98). This is not a bug.
It is the known perception-distortion trade-off~\cite{blau2018perception} inside one network: the adversarial and identity losses act only on the last cycle and push it toward a sharper, more realistic image, which lowers pixel fidelity a little.
We look at this next.

\paragraph{Parity with the full GFPGAN}
The comparison above uses only deploy-compatible from-scratch baselines, since no $512\times512$ prior-guided model compiles on our NPU.
For context, we also ran the full official GFPGAN, with its frozen StyleGAN2 prior at its native resolution, on the same inputs.
CFRNet at $k\!=\!3$ gives about the same visual quality as this much larger model (see also Section~\ref{subsec:direct}), while using a small fraction of its parameters and compute and, unlike it, running on the Hi3402.
We report this as a qualitative observation, not a benchmark number, because the two models work at different native resolutions.
Still, reaching the quality of GFPGAN's pretrained-prior pipeline with a 2.0\,M-parameter standard-CNN restorer is a useful result for deployment.
\subsection{Qualitative Results}

\begin{figure*}[t]
\centering
\includegraphics[width=0.92\linewidth]{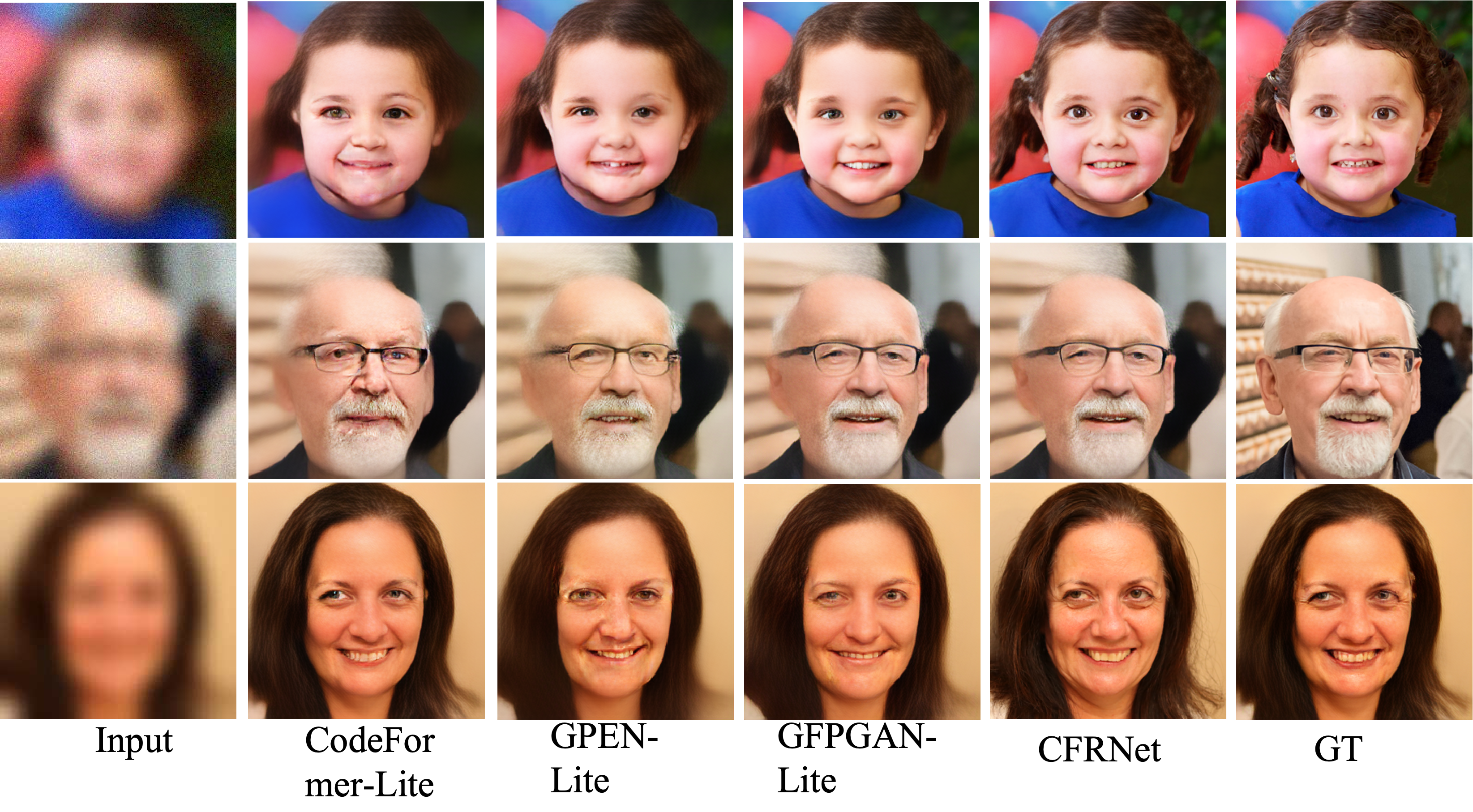}
\caption{Qualitative comparison on the FFHQ-256 test set. Left to right: degraded input, CodeFormer-Lite, GPEN-Lite, GFPGAN-Lite, CFRNet ($k\!=\!3$), ground truth.
All baselines are trained from scratch under the same protocol.
Across a child, an older man with glasses, and an adult, CFRNet gives comparable or finer detail in the eyes, hair, and skin with 5 to 8 times fewer parameters, and with fewer mid-face artifacts than the Lite baselines.
The glasses frames and beard in the second row, which are hard for small restorers, stay sharp and free of ringing.}
\label{fig:qual_main}
\end{figure*}

\begin{figure*}[t]
\centering
\includegraphics[width=0.92\linewidth]{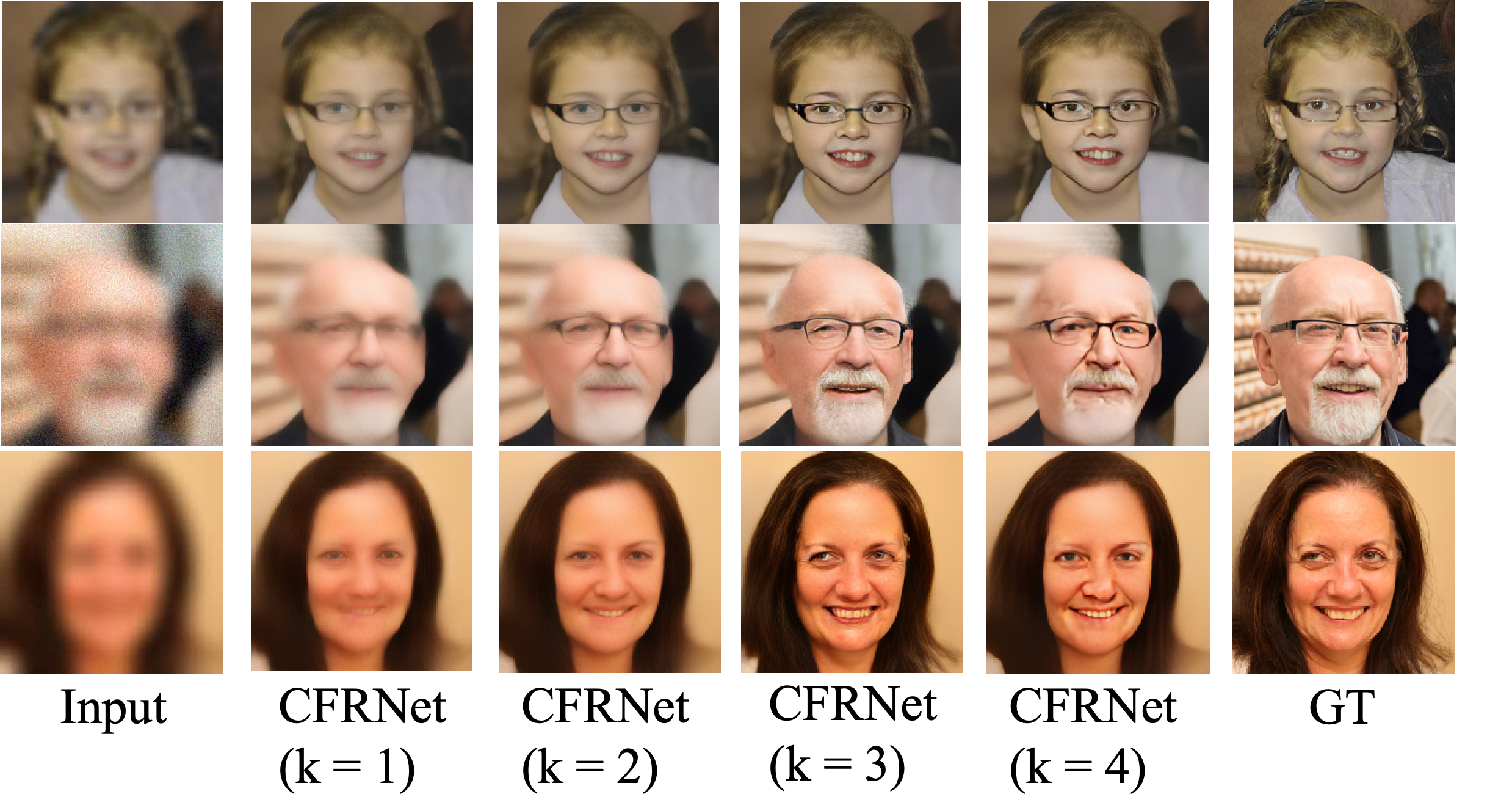}
\caption{Cycle-by-cycle output of one trained CFRNet.
Left to right: degraded input, $k\!=\!1$, $k\!=\!2$, $k\!=\!3$, $k\!=\!4$, ground truth. Every cycle reuses the same 2.0\,M weights.
Early cycles mostly denoise; later cycles sharpen the face and restore detail in the eyes, mouth, and hair.
The $k\!=\!3$ and $k\!=\!4$ outputs look almost the same, which is the fixed-point behavior from CCFP and matches Tables~\ref{tab:cycle_ablation} and~\ref{tab:fixedpoint}.}
\label{fig:cycle_progress}
\end{figure*}

\begin{figure}[t]
\centering
\includegraphics[width=\linewidth]{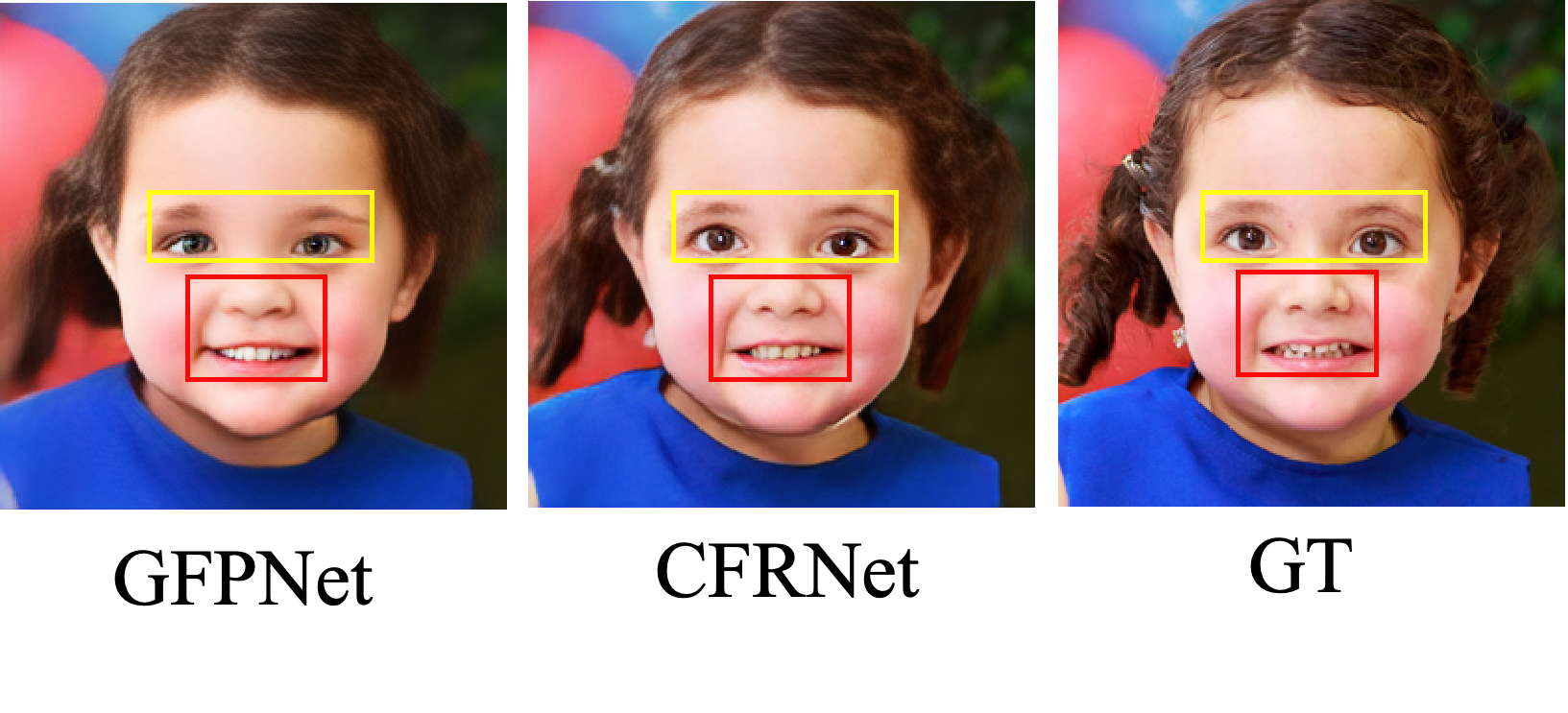}
\caption{Component-level comparison with zoom-ins.
Yellow box: eye region; red box: mouth region. Left to right: GFPGAN-Lite (labeled GFPNet in the panel), CFRNet ($k\!=\!3$), ground truth.
CFRNet restores the eye and the teeth and mouth closer to the ground truth, while the small GFPGAN-Lite baseline tends to over-smooth the iris and blur the teeth.
The mouth result reflects the added nose and mouth supervision (Section~\ref{subsec:component});
the matching numbers are in Table~\ref{tab:ablation_component}.}
\label{fig:roi}
\end{figure}

Fig.~\ref{fig:qual_main} compares CFRNet ($k\!=\!3$) with the retrained Lite baselines.
Fig.~\ref{fig:cycle_progress} shows the cycle-by-cycle output of one model and makes the fixed-point behavior easy to see: the output stops changing once it reaches a natural face.
Fig.~\ref{fig:roi} zooms into the eye and mouth, where small baselines fail most often, and shows that CFRNet follows the ground truth more closely.
\subsection{Effect of the Cycle Count}
\label{subsec:cycle_ablation}

Table~\ref{tab:cycle_ablation} lists the metrics and the latency on both platforms across cycles.
\begin{table}[t]
\centering
\caption{Effect of the cycle count for one trained CFRNet. PSNR is best at $k\!=\!2$, LPIPS is best at $k\!=\!3$, and both drop only a little at $k\!=\!4$.
The model was trained with $k_{\mathrm{train}}=3$, so $k\!=\!4$ is one pass beyond training, yet the change is small.
Server latency is V100 FP32; board latency is Hi3402 INT8.}
\label{tab:cycle_ablation}
\begin{tabular}{cccccc}
\toprule
$k$ & PSNR $\uparrow$ & SSIM $\uparrow$ & LPIPS $\downarrow$ & Server (ms) & Board (ms) \\
\midrule
1 & 23.54          & 0.673           & 0.360           & 2.7 & 23 \\
2 & \textbf{23.82} & \textbf{0.685}  & 0.300           & 4.9 & 46 \\
3 & 23.10          & 
0.662           & \textbf{0.250}  & 7.3 & 69 \\
4 & 22.98          & 0.668           & 0.262           & 9.6 & 92 \\
\bottomrule
\end{tabular}
\end{table}

\paragraph{Why the curves differ}
The adversarial and identity losses act only on the final cycle $x^{(k)}$.
The earlier cycles see only the reconstruction terms. So the network spreads the work out: the first two cycles mostly denoise, which helps PSNR, and the third cycle adds the adversarial and identity detail, which helps LPIPS at a small cost in PSNR.
This is the perception-distortion trade-off~\cite{blau2018perception} seen inside one model.

\paragraph{How to choose $k$}
The best cycle count depends on the use case.
Use $k\!=\!3$ for the best perceptual quality, for example photo and video apps where a person looks at the output.
Use $k\!=\!2$ for the best PSNR and SSIM, for example when the restored image feeds a later pixel-level step.
Use $k\!=\!1$ for the lowest latency on a very tight budget.
Because $k$ is chosen at inference, no retraining or weight reloading is needed to move along this curve.
\paragraph{Fixed-point check}
Since the model was trained with $k_{\mathrm{train}}=3$, the $k\!=\!4$ result runs $G$ once more than training did.
The output barely moves: LPIPS rises by only 0.012 and PSNR drops by only 0.12\,dB.
This means $x^{(3)}$ is close to a fixed point of $G$.
A single-pass model behaves differently and drifts at $k>k_{\mathrm{train}}$ (the red curve in Fig.~\ref{fig:dynamics}), as the next section shows.
\subsection{Ablation: CCFP Losses}
\label{subsec:ablation_ccfp}

Table~\ref{tab:ablation_ccfp} isolates each CCFP loss. All rows use the same generator and run at $k\!=\!3$;
they differ only in the training loss.

\begin{table}[t]
\centering
\caption{Ablation of the CCFP losses at $k\!=\!3$.
``Single-pass'' is standard one-pass training reused $k\!=\!3$ times at test time, the setting used by earlier recurrent work~\cite{lee2019recurrent}.
Each row adds the listed loss to the row above.}
\label{tab:ablation_ccfp}
\begin{tabular}{lccc}
\toprule
Training setting & PSNR $\uparrow$ & LPIPS $\downarrow$ & SSIM $\uparrow$ \\
\midrule
Single-pass (heuristic $k\!=\!3$)                              & 22.62          & 0.330          & 0.636 \\
\quad $+$ progressive $\mathcal{L}_{\mathrm{prog}}$            & 22.88        
  & 0.292          & 0.650 \\
\quad $+$ idempotence $\mathcal{L}_{\mathrm{fp}}$              & 23.02          & 0.271          & 0.657 \\
\quad $+$ cycle $\mathcal{L}_{\mathrm{cyc}}$ (full CCFP)       & \textbf{23.10} & \textbf{0.250} & \textbf{0.662} \\
\midrule
\textit{gain over single-pass} & $+$0.48 & $-$0.080 & $+$0.026 \\
\bottomrule
\end{tabular}
\end{table}

Each loss helps.
Progressive supervision gives the largest single LPIPS drop ($-0.038$) by removing the train-test mismatch.
The idempotence loss adds another $-0.021$ and makes the cycles stable, which we check below.
The cycle loss adds another $-0.021$, mostly by improving results on harder degradations.
\paragraph{Checking the fixed-point property}
\label{par:fp_property}
To check that CCFP gives a contractive operator, we measure the per-cycle change $\Delta_k=\|G^{(k+1)}(x)-G^{(k)}(x)\|_2$ on the test set.
\begin{table}[t]
\centering
\caption{Per-cycle change $\Delta_k$ (mean $L_2$ over the test set; smaller at large $k$ is more fixed-point-like).
CCFP converges much more tightly than the single-pass model.}
\label{tab:fixedpoint}
\begin{tabular}{lccc}
\toprule
Setting & $\Delta_2$ & $\Delta_3$ & $\Delta_4$ \\
\midrule
Single-pass training & 0.071 & 0.058 & 0.052 \\
Full CCFP            & \textbf{0.043} & \textbf{0.022} & \textbf{0.018} \\
\bottomrule
\end{tabular}
\end{table}

The single-pass model still changes a lot at $k\!=\!4$, while CCFP cuts $\Delta_4$ by about 65\%.
This shows that CCFP makes $G$ settle near $\mathcal{M}_y$, which matches Fig.~\ref{fig:dynamics}.
\subsection{Ablation: Component Losses and the Nose Region}

\begin{table}[t]
\centering
\caption{Effect of the face-component discriminators (CFRNet at $k\!=\!3$, full CCFP).
Each row adds the listed supervision to the row above.}
\label{tab:ablation_component}
\begin{tabular}{lccc}
\toprule
Training objective                            & PSNR $\uparrow$ & SSIM $\uparrow$ & LPIPS $\downarrow$ \\
\midrule
No component losses                           & 22.90          & 0.655          & 0.285 
\\
\quad $+$ eye/mouth (GFPGAN-style~\cite{wang2021gfpgan}) & 23.02 & 0.659 & 0.266 \\
\quad $+$ eye/mouth/nose (full)              & \textbf{23.10} & \textbf{0.662} & \textbf{0.250} \\
\bottomrule
\end{tabular}
\end{table}

Adding the nose region gives another $-0.016$ LPIPS over the eye-and-mouth setting.
In the images (Fig.~\ref{fig:qual_main} and Fig.~\ref{fig:roi}) the nose region removes ringing around the nostrils and the nasal bridge, a common problem for small CNN restorers.
\subsection{Deployment on the Consumer NPU}

\begin{table}[t]
\centering
\caption{Deployment on the HiSilicon Hi3402 NPU.
The INT8 toolchain compiles CFRNet end to end but cannot compile the operator sets that the Lite baselines still use, such as the codebook lookup in CodeFormer-Lite and the style modulation in GPEN-Lite.}
\label{tab:deployment}
\resizebox{\linewidth}{!}{%
\begin{tabular}{lcccc}
\toprule
Item & GFPGAN-Lite & GPEN-Lite & CodeFormer-Lite & CFRNet \\
\midrule
Input size                  & $256{\times}256$ & $256{\times}256$ & $256{\times}256$ & $256{\times}256$ \\
Params (M) $\downarrow$     & $\sim$17 & $\sim$15 & $\sim$10 & \textbf{2.0} \\
MACs per pass (G)           & $\sim$23 & 
$\sim$19 & $\sim$11 & \textbf{8.7} \\
INT8 quantization           & partial$^*$ & no$^\dagger$ & no$^\ddagger$ & \textbf{yes} \\
Hi3402 compilable           & no & no & no & \textbf{yes} \\
Board latency (ms)          & infeasible & infeasible & infeasible & \textbf{23 ($k\!=\!1$) / 69 ($k\!=\!3$)} \\
\bottomrule
\end{tabular}}\\
\vspace{2pt}
{\footnotesize $^*$ U-Net skip-concatenation shapes need an FP16 fallback.
$^\dagger$ Style-modulation ops are not in the Hi3402 INT8 op set.
$^\ddagger$ Codebook nearest-neighbor lookup is not a standard CNN op.}
\end{table}

On the Hi3402, CFRNet finishes a 3-cycle restoration in about 69\,ms at $256\times256$, or about 23\,ms per cycle.
For comparison, the same model runs in about 2.4\,ms per cycle on a V100 in FP32 (Table~\ref{tab:cycle_ablation}).
The gap is expected, since the V100 is a large datacenter GPU and the Hi3402 is a small embedded NPU;
both numbers describe the same model at the two ends of the deployment range.
The board latency leaves room for face detection, alignment, and rendering inside a real-time budget on a device that cannot run any of the baselines we tested.
\subsection{A Simpler Variant: A Plain Iterated CNN}
\label{subsec:direct}

CCFP gives our best perceptual quality, but it has a cost at training time.
The multi-cycle unroll plus the idempotence and cycle passes make training take about twice as long as standard single-pass training.
For teams that want the simplest possible pipeline, we asked whether the main idea, that running a small restorer again moves it toward a natural face, still works in a much simpler setting.
So we trained a separate, plain CNN, which we call CFRNet-D, with the standard single-pass recipe (no progressive unroll, no idempotence, no cycle loss), and at test time we just fed its output back in. We write direct-$j$ for $j$ passes.
This model is very easy to train and to deploy: one static graph, one operator set, and refinement by simply running the same graph again, at about 23\,ms per pass on the board.
\begin{figure}[t]
\centering
\includegraphics[width=\linewidth]{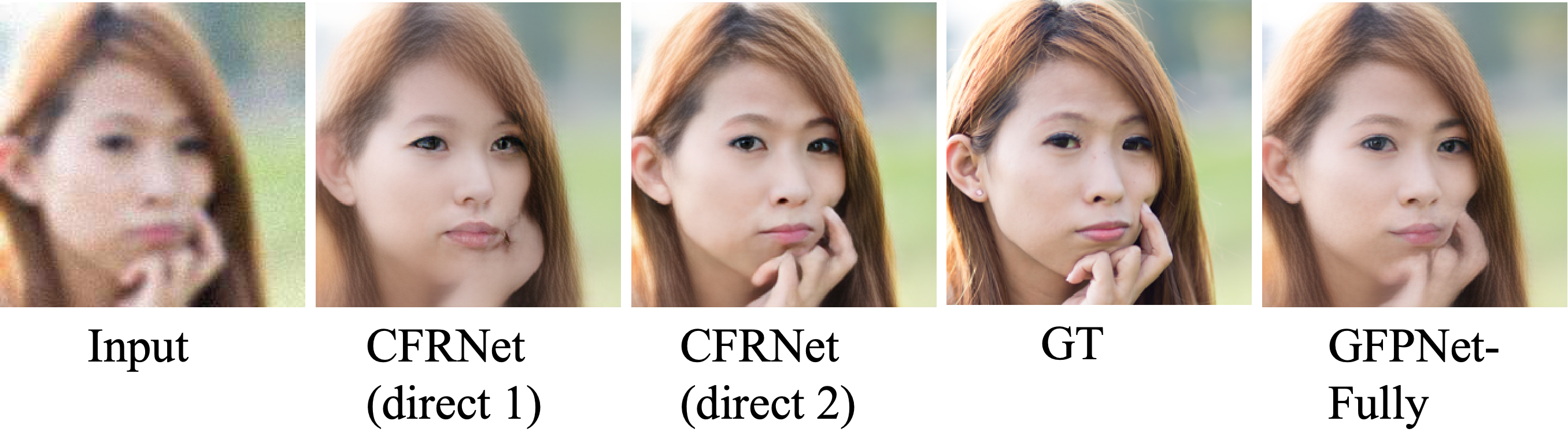}
\caption{The plain variant on a hard real-world input (heavy haze, low contrast, far from the FFHQ training data).
Left to right: degraded input, CFRNet-D direct-1 (one pass), CFRNet-D direct-2 (the same plain network run twice), ground truth, and GFPGAN-Lite (labeled GFPNet).
One pass already removes most of the degradation. Because the second pass sees a cleaner input, it recovers more detail and reaches about the quality of the full CCFP model at $k\!=\!3$, and it is better than the small GFPGAN-Lite baseline on this input.
On such inputs we also found CFRNet-D direct-2 to be about the same as the full official GFPGAN, here with a network small enough to run at 23\,ms per pass on the board.}
\label{fig:direct}
\end{figure}

Fig.~\ref{fig:direct} shows a hard real-world example.
One pass (direct-1) removes most of the degradation, and the second pass (direct-2) sees a cleaner input and recovers more detail, reaching about the quality of the full CCFP model at $k\!=\!3$ and beating our GFPGAN-Lite baseline on this input.
We do not present CFRNet-D as a replacement for CCFP.
CCFP is the principled, best-quality model with a built-in fixed-point property and a no-retrain quality knob.
CFRNet-D simply shows that the iterative idea is robust enough to work even with a very plain, easy-to-ship network.
Two points follow. First, parameter count and architectural complexity are not the only things that set restoration quality;
how inference is structured can matter as much as model size.
Second, for consumer products where integration cost often matters more than training cost, a plain iterated CNN can be a good practical choice.
We think inference-time structure deserves more attention when restoration models go to limited hardware.
\subsection{Real-Time In-Car Deployment}
\label{subsec:inwild}

\begin{figure}[t]
\centering
\includegraphics[width=0.8\linewidth]{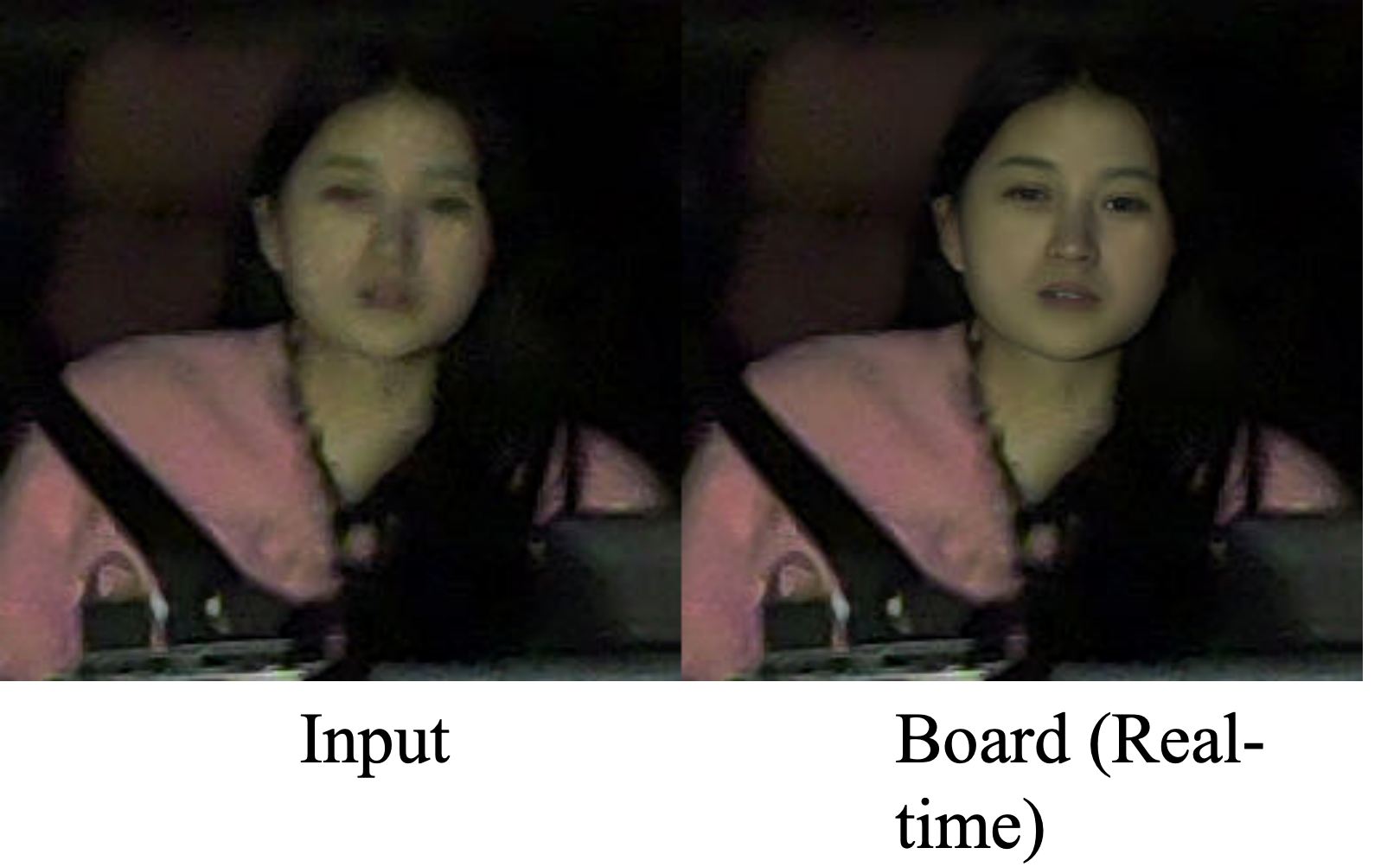}
\caption{Real-time output on the embedded board in a driver-monitoring setting.
Left: raw on-device input (low light, motion, strong shadows, far from the FFHQ training data). Right: the board's real-time output.
The model restores the face and skin tone directly from the on-device stream inside the per-frame budget, with no offline pre- or post-processing.
This is the setting we ship.}
\label{fig:inwild}
\end{figure}

Fig.~\ref{fig:inwild} shows a frame captured and restored in real time on the board in a driver-monitoring setting, which is low light and motion-heavy and far from the FFHQ training data.
The model restores the face and skin tone directly from the on-device stream inside the per-frame budget, with no offline steps.
This in-the-wild result on inputs that differ from the training data is helped by the re-degradation cycle loss and by the stability that the fixed-point training gives.
% ============================================================
\section{Discussion and Limitations}
% ============================================================

\paragraph{What CCFP does}
A single-pass $G$ is trained only on degraded inputs $\mathcal{X}$.
Its outputs $G(\mathcal{X})$ are close to, but not on, the natural-face set $\mathcal{M}_y$.
Running $G$ again on $G(x)$ is therefore out of distribution, and the result is not controlled, so it can over-sharpen, drift in color, or oscillate (the red curve in Fig.~\ref{fig:dynamics}).
CCFP fixes this by training $G$ on its own outputs (progressive supervision), by penalizing further change at convergence (idempotence), and by encoding the degradation process (re-degradation cycle).
\paragraph{Why PSNR peaks at $k\!=\!2$}
This is the perception-distortion trade-off~\cite{blau2018perception} inside one network.
The adversarial and identity losses act only on the last cycle, so the network uses that cycle to add realistic detail, which helps LPIPS but slightly lowers PSNR.
For deployment this is fine, because the task is perceptual and a person looks at the output.
The choice between $k\!=\!2$ and $k\!=\!3$ gives the user a simple, no-retrain knob along the trade-off.
\paragraph{Limitations}
First, latency grows linearly with $k$. On the Hi3402 $k\!=\!3$ is fast enough, but a tighter chip may prefer $k\!=\!1$ or $k\!=\!2$, with the quality shown in Section~\ref{subsec:cycle_ablation}.
Second, the re-degradation cycle loss assumes the deployment degradation is similar to the training degradation $D$.
A very different degradation, such as extreme low light or motion blur that $D$ does not cover, may not be handled, although the in-car result in Section~\ref{subsec:inwild} is encouraging.
Third, CCFP training takes about twice as long as single-pass training, because each step runs $G$ several times plus the idempotence and cycle passes.
This is a one-time cost and does not affect inference, and the plain CFRNet-D variant (Section~\ref{subsec:direct}) trades a small quality margin for standard training cost.
Fourth, our numbers are against from-scratch Lite reimplementations of GFPGAN, GPEN, and CodeFormer, which is the right setting for on-device use;
the parity note against the full official GFPGAN is a qualitative observation at a different native resolution, not a benchmark number.
% ============================================================
\section{Conclusion}
% ============================================================
We presented CFRNet, a small blind face restoration model for consumer NPUs, and CCFP, its training method.
CCFP closes a gap in iterative restoration: instead of training a network for one mapping and reusing it by hand, we train it as a fixed-point operator on the set of natural faces, using progressive supervision, an idempotence loss, and a re-degradation cycle loss.
With a 2.0\,M-parameter NPU-friendly generator and component supervision that adds a nose region, CFRNet gives the best perceptual score among comparable from-scratch baselines on a 300-image test set, gives the best PSNR and SSIM at $k\!=\!2$, and is the only model that runs on the target NPU.
It also reaches about the quality of the full official GFPGAN.
The cycle count is a simple, no-retrain knob between pixel fidelity ($k\!=\!2$) and perceptual quality ($k\!=\!3$).
Finally, the same idea works with a plain, easy-to-ship CNN that reaches $k\!=\!3$ quality on hard inputs, and the model runs in real time on an in-car driver-monitoring board.
Future work includes extending CCFP to $512\times512$ deployments, testing on larger real-world sets, and learning a per-image stopping rule for the cycles.
% ============================================================
\section*{Acknowledgment}
The authors thank the reviewers for their comments. The source code will be made publicly available upon acceptance of this manuscript.
The authors also acknowledge the use of large language models (Google Gemini) strictly for English language editing and proofreading during the preparation of this manuscript.
\bibliographystyle{IEEEtran}
\bibliography{refs}

@inproceedings{wang2021gfpgan,
  title={Towards real-world blind face restoration with generative facial prior},
  author={Wang, Xintao and Li, Yu and Zhang, Honglun and Shan, Ying},
  booktitle={Proceedings of the IEEE/CVF Conference on Computer Vision and Pattern Recognition (CVPR)},
  pages={9168--9178},
  year={2021}
}

@inproceedings{yang2021gpen,
  title={GAN prior embedded network for blind face restoration in the wild},
  author={Yang, Tao and Ren, Peiran and Xie, Xuansong and Zhang, Lei},
  booktitle={Proceedings of the IEEE/CVF Conference on Computer Vision and Pattern Recognition (CVPR)},
  pages={672--681},
  year={2021}
}

@inproceedings{wang2022restoreformer,
  title={RestoreFormer: High-quality blind face restoration from undegraded key-value pairs},
  author={Wang, Zhouxia and Zhang, Jiawei and Chen, Runjian and Wang, Wenping and Luo, Ping},
  booktitle={Proceedings of the IEEE/CVF Conference on Computer Vision and Pattern Recognition (CVPR)},
  pages={17512--17521},
  year={2022}
}

@inproceedings{zhou2022codeformer,
  title={Towards robust blind face restoration with codebook lookup transformer},
  author={Zhou, Shangchen and Chan, Kelvin C.K. and Li, Chongyi and Loy, Chen Change},
  booktitle={Advances in Neural Information Processing Systems (NeurIPS)},
  volume={35},
  pages={30599--30611},
  year={2022}
}

@inproceedings{sandler2018mobilenetv2,
  title={MobileNetV2: Inverted residuals and linear bottlenecks},
  author={Sandler, Mark and Howard, Andrew and Zhu, Menglong and Zhmoginov, Andrey and Chen, Liang-Chieh},
  booktitle={Proceedings of the IEEE Conference on Computer Vision and Pattern Recognition (CVPR)},
  pages={4510--4520},
  year={2018}
}

@inproceedings{howard2019mobilenetv3,
  title={Searching for MobileNetV3},
  author={Howard, Andrew and Sandler, Mark and Chu, Grace and Chen, Liang-Chieh and Chen, Bo and Tan, Mingxing and Wang, Weijun and Zhu, Yukun and Pang, Ruoming and Vasudevan, Vijay and Le, Quoc V. and Adam, Hartwig},
  booktitle={Proceedings of the IEEE/CVF International Conference on Computer Vision (ICCV)},
  pages={1314--1324},
  year={2019}
}

@inproceedings{liu2020rfdn,
  title={Residual feature distillation network for lightweight image super-resolution},
  author={Liu, Jie and Tang, Jie and Wu, Gangshan},
  booktitle={European Conference on Computer Vision (ECCV) Workshops},
  pages={41--55},
  year={2020}
}

@inproceedings{haris2018dbpn,
  title={Deep back-projection networks for super-resolution},
  author={Haris, Muhammad and Shakhnarovich, Gregory and Ukita, Norimichi},
  booktitle={Proceedings of the IEEE Conference on Computer Vision and Pattern Recognition (CVPR)},
  pages={1664--1673},
  year={2018}
}

@inproceedings{ma2020dic,
  title={Deep face super-resolution with iterative collaboration between attentive recovery and landmark estimation},
  author={Ma, Cheng and Jiang, Zhenyu and Rao, Yongming and Lu, Jiwen and Zhou, Jie},
  booktitle={Proceedings of the IEEE/CVF Conference on Computer Vision and Pattern Recognition (CVPR)},
  pages={5569--5578},
  year={2020}
}

@inproceedings{lee2019recurrent,
  title={Feedback network for image super-resolution},
  author={Li, Zhen and Yang, Jinglei and Liu, Zheng and Yang, Xiaomin and Jeon, Gwanggil and Wu, Wei},
  booktitle={Proceedings of the IEEE/CVF Conference on Computer Vision and Pattern Recognition (CVPR)},
  pages={3867--3876},
  year={2019}
}

@inproceedings{ledig2017photo,
  title={Photo-realistic single image super-resolution using a generative adversarial network},
  author={Ledig, Christian and Theis, Lucas and Husz{\'a}r, Ferenc and Caballero, Jose and Cunningham, Andrew and Acosta, Alejandro and Aitken, Andrew and Tejani, Alykhan and Totz, Johannes and Wang, Zehan and Shi, Wenzhe},
  booktitle={Proceedings of the IEEE Conference on Computer Vision and Pattern Recognition (CVPR)},
  pages={4681--4690},
  year={2017}
}

@inproceedings{wang2018esrgan,
  title={ESRGAN: Enhanced super-resolution generative adversarial networks},
  author={Wang, Xintao and Yu, Ke and Wu, Shixiang and Gu, Jinjin and Liu, Yihao and Dong, Chao and Qiao, Yu and Loy, Chen Change},
  booktitle={European Conference on Computer Vision (ECCV) Workshops},
  pages={63--79},
  year={2018}
}

@inproceedings{bulat2018superfan,
  title={Super-FAN: Integrated facial landmark localization and super-resolution of real-world low resolution faces in arbitrary poses with GANs},
  author={Bulat, Adrian and Tzimiropoulos, Georgios},
  booktitle={Proceedings of the IEEE Conference on Computer Vision and Pattern Recognition (CVPR)},
  pages={109--117},
  year={2018}
}

@inproceedings{chen2018fsrnet,
  title={FSRNet: End-to-end learning face super-resolution with facial priors},
  author={Chen, Yu and Tai, Ying and Liu, Xiaoming and Shen, Chunhua and Yang, Jian},
  booktitle={Proceedings of the IEEE Conference on Computer Vision and Pattern Recognition (CVPR)},
  pages={2492--2501},
  year={2018}
}

@inproceedings{li2020dfdnet,
  title={Blind face restoration via deep multi-scale component dictionaries},
  author={Li, Xiaoming and Chen, Chaofeng and Zhou, Shangchen and Lin, Xianhui and Zuo, Wangmeng and Zhang, Lei},
  booktitle={European Conference on Computer Vision (ECCV)},
  pages={399--415},
  year={2020}
}

@article{saharia2022sr3,
  title={Image super-resolution via iterative refinement},
  author={Saharia, Chitwan and Ho, Jonathan and Chan, William and Salimans, Tim and Fleet, David J. and Norouzi, Mohammad},
  journal={IEEE Transactions on Pattern Analysis and Machine Intelligence (TPAMI)},
  volume={45},
  number={4},
  pages={4713--4726},
  year={2023}
}

@article{delbracio2023indi,
  title={Inversion by direct iteration: An alternative to denoising diffusion for image restoration},
  author={Delbracio, Mauricio and Milanfar, Peyman},
  journal={Transactions on Machine Learning Research (TMLR)},
  year={2023}
}

@inproceedings{bai2019deq,
  title={Deep equilibrium models},
  author={Bai, Shaojie and Kolter, J. Zico and Koltun, Vladlen},
  booktitle={Advances in Neural Information Processing Systems (NeurIPS)},
  year={2019}
}

@inproceedings{song2023consistency,
  title={Consistency models},
  author={Song, Yang and Dhariwal, Prafulla and Chen, Mark and Sutskever, Ilya},
  booktitle={International Conference on Machine Learning (ICML)},
  pages={32211--32252},
  year={2023}
}

@inproceedings{zhu2017cyclegan,
  title={Unpaired image-to-image translation using cycle-consistent adversarial networks},
  author={Zhu, Jun-Yan and Park, Taesung and Isola, Phillip and Efros, Alexei A.},
  booktitle={Proceedings of the IEEE International Conference on Computer Vision (ICCV)},
  pages={2223--2232},
  year={2017}
}

@inproceedings{refstar2025,
  title={RefSTAR: Blind face image restoration with reference selection, transfer, and reconstruction},
  author={Yin, Zhicun and Chen, Junjie and Liu, Ming and Wang, Zhixin and Li, Fan and Pei, Renjing and Li, Xiaoming and Lau, Rynson W.H. and Zuo, Wangmeng},
  booktitle={Proceedings of the AAAI Conference on Artificial Intelligence},
  volume={40},
  number={14},
  pages={12053--12062},
  year={2026}
}

@inproceedings{kuai2024unsupervised,
  title={Towards unsupervised blind face restoration using diffusion prior},
  author={Kuai, Tianshu and Honari, Sina and Gilitschenski, Igor and Levinshtein, Alex},
  booktitle={Proceedings of the IEEE/CVF Winter Conference on Applications of Computer Vision (WACV)},
  pages={1839--1849},
  year={2025}
}

@inproceedings{blau2018perception,
  title={The perception-distortion tradeoff},
  author={Blau, Yochai and Michaeli, Tomer},
  booktitle={Proceedings of the IEEE Conference on Computer Vision and Pattern Recognition (CVPR)},
  pages={6228--6237},
  year={2018}
}

@inproceedings{karras2019stylegan,
  title={A style-based generator architecture for generative adversarial networks},
  author={Karras, Tero and Laine, Samuli and Aila, Timo},
  booktitle={Proceedings of the IEEE/CVF Conference on Computer Vision and Pattern Recognition (CVPR)},
  pages={4401--4410},
  year={2019}
}

@inproceedings{zhang2018lpips,
  title={The unreasonable effectiveness of deep features as a perceptual metric},
  author={Zhang, Richard and Isola, Phillip and Efros, Alexei A. and Shechtman, Eli and Wang, Oliver},
  booktitle={Proceedings of the IEEE Conference on Computer Vision and Pattern Recognition (CVPR)},
  pages={586--595},
  year={2018}
}

@inproceedings{goodfellow2014gan,
  title={Generative adversarial nets},
  author={Goodfellow, Ian J. and Pouget-Abadie, Jean and Mirza, Mehdi and Xu, Bing and Warde-Farley, David and Ozair, Sherjil and Courville, Aaron and Bengio, Yoshua},
  booktitle={Advances in Neural Information Processing Systems (NeurIPS)},
  pages={2672--2680},
  year={2014}
}

@inproceedings{johnson2016perceptual,
  title={Perceptual losses for real-time style transfer and super-resolution},
  author={Johnson, Justin and Alahi, Alexandre and Fei-Fei, Li},
  booktitle={European Conference on Computer Vision (ECCV)},
  pages={694--711},
  year={2016}
}

@inproceedings{isola2017pix2pix,
  title={Image-to-image translation with conditional adversarial networks},
  author={Isola, Phillip and Zhu, Jun-Yan and Zhou, Tinghui and Efros, Alexei A.},
  booktitle={Proceedings of the IEEE Conference on Computer Vision and Pattern Recognition (CVPR)},
  pages={1125--1134},
  year={2017}
}

@inproceedings{miyato2018spectral,
  title={Spectral normalization for generative adversarial networks},
  author={Miyato, Takeru and Kataoka, Toshiki and Koyama, Masanori and Yoshida, Yuichi},
  booktitle={International Conference on Learning Representations (ICLR)},
  year={2018}
}

@inproceedings{deng2019arcface,
  title={ArcFace: Additive angular margin loss for deep face recognition},
  author={Deng, Jiankang and Guo, Jia and Xue, Niannan and Zafeiriou, Stefanos},
  booktitle={Proceedings of the IEEE/CVF Conference on Computer Vision and Pattern Recognition (CVPR)},
  pages={4690--4699},
  year={2019}
}

@article{dong2016srcnn,
  title={Image super-resolution using deep convolutional networks},
  author={Dong, Chao and Loy, Chen Change and He, Kaiming and Tang, Xiaoou},
  journal={IEEE Transactions on Pattern Analysis and Machine Intelligence (TPAMI)},
  volume={38},
  number={2},
  pages={295--307},
  year={2016}
}

@inproceedings{lim2017edsr,
  title={Enhanced deep residual networks for single image super-resolution},
  author={Lim, Bee and Son, Sanghyun and Kim, Heewon and Nah, Seungjun and Lee, Kyoung Mu},
  booktitle={Proceedings of the IEEE Conference on Computer Vision and Pattern Recognition (CVPR) Workshops},
  pages={136--144},
  year={2017}
}

@inproceedings{zhang2018rcan,
  title={Image super-resolution using very deep residual channel attention networks},
  author={Zhang, Yulun and Li, Kunpeng and Li, Kai and Wang, Lichen and Zhong, Bineng and Fu, Yun},
  booktitle={European Conference on Computer Vision (ECCV)},
  pages={286--301},
  year={2018}
}

@inproceedings{liang2021swinir,
  title={SwinIR: Image restoration using Swin transformer},
  author={Liang, Jingyun and Cao, Jiezhang and Sun, Guolei and Zhang, Kai and Van Gool, Luc and Timofte, Radu},
  booktitle={Proceedings of the IEEE/CVF International Conference on Computer Vision (ICCV) Workshops},
  pages={1833--1844},
  year={2021}
}

@inproceedings{wang2021realesrgan,
  title={Real-ESRGAN: Training real-world blind super-resolution with pure synthetic data},
  author={Wang, Xintao and Xie, Liangbin and Dong, Chao and Shan, Ying},
  booktitle={Proceedings of the IEEE/CVF International Conference on Computer Vision (ICCV) Workshops},
  pages={1905--1914},
  year={2021}
}

@inproceedings{gu2022vqfr,
  title={VQFR: Blind face restoration with vector-quantized dictionary and parallel decoder},
  author={Gu, Yuchao and Wang, Xintao and Xie, Liangbin and Dong, Chao and Li, Gen and Shan, Ying and Cheng, Ming-Ming},
  booktitle={European Conference on Computer Vision (ECCV)},
  pages={126--143},
  year={2022}
}

@article{yue2024difface,
  title={DifFace: Blind face restoration with diffused error contraction},
  author={Yue, Zongsheng and Loy, Chen Change},
  journal={IEEE Transactions on Pattern Analysis and Machine Intelligence (TPAMI)},
  volume={46},
  number={12},
  pages={9991--10004},
  year={2024}
}

@inproceedings{jacob2018quantization,
  title={Quantization and training of neural networks for efficient integer-arithmetic-only inference},
  author={Jacob, Benoit and Kligys, Skirmantas and Chen, Bo and Zhu, Menglong and Tang, Matthew and Howard, Andrew and Adam, Hartwig and Kalenichenko, Dmitry},
  booktitle={Proceedings of the IEEE Conference on Computer Vision and Pattern Recognition (CVPR)},
  pages={2704--2713},
  year={2018}
}

@inproceedings{tan2019efficientnet,
  title={EfficientNet: Rethinking model scaling for convolutional neural networks},
  author={Tan, Mingxing and Le, Quoc V.},
  booktitle={International Conference on Machine Learning (ICML)},
  pages={6105--6114},
  year={2019}
}

@inproceedings{reddy2017drowsiness,
  title={Real-time driver drowsiness detection for embedded system using model compression of deep neural networks},
  author={Reddy, Bhargava and Kim, Ye-Hoon and Yun, Sojung and Seo, Chanwon and Jang, Junik},
  booktitle={Proceedings of the IEEE Conference on Computer Vision and Pattern Recognition (CVPR) Workshops},
  pages={438--445},
  year={2017}
}

@article{yu2019drowsiness,
  title={Driver drowsiness detection using condition-adaptive representation learning framework},
  author={Yu, Jongmin and Park, Sangwook and Lee, Sangwook and Jeon, Moongu},
  journal={IEEE Transactions on Intelligent Transportation Systems},
  volume={20},
  number={11},
  pages={4206--4218},
  year={2019}
}

% ============================================================
% Author Biographies
% ============================================================

\begin{IEEEbiography}[{\includegraphics[width=1in,height=1.25in,clip,keepaspectratio]{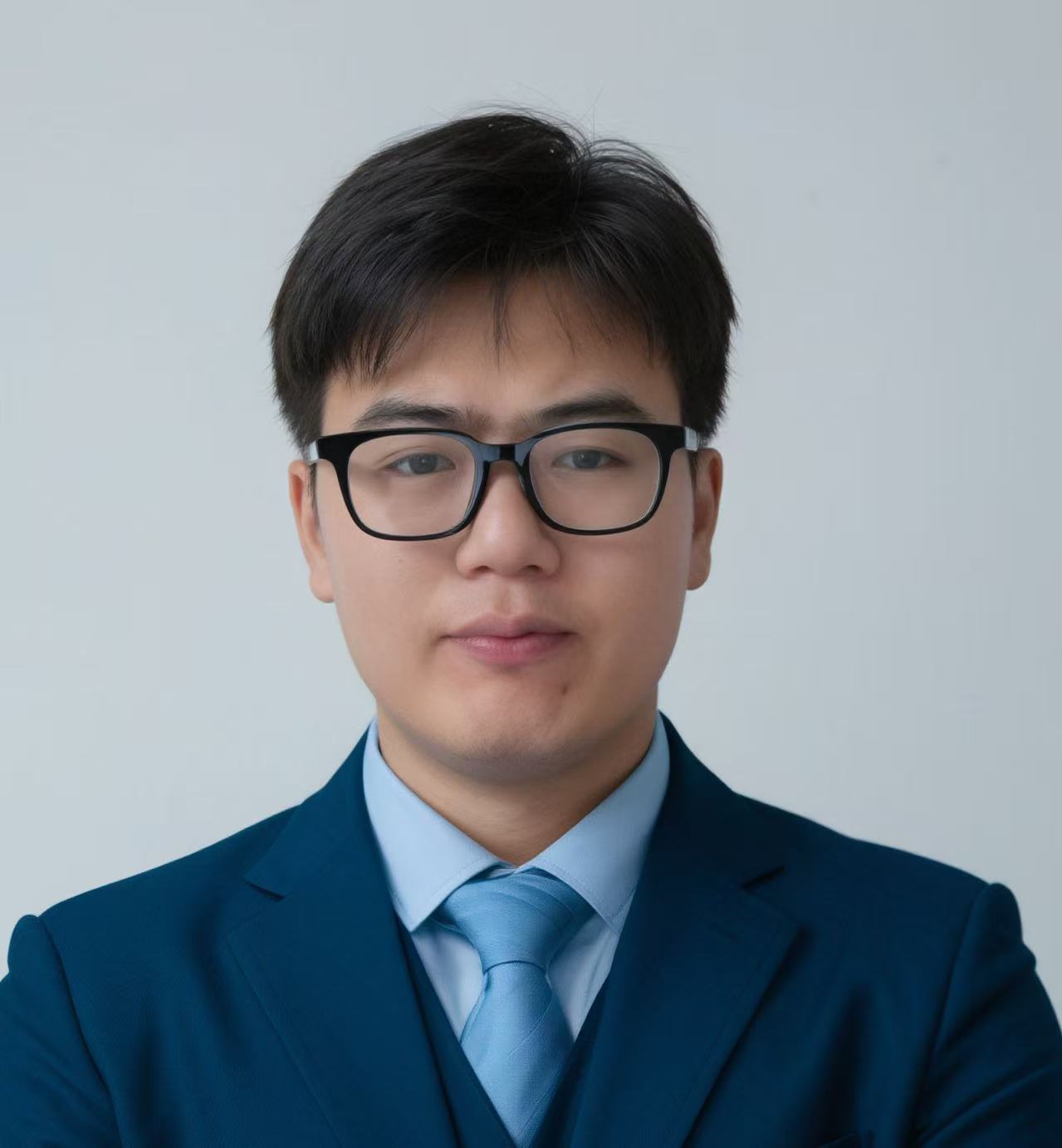}}]{Fuchen Li}
Fuchen Li is currently pursuing the Second M.S.
degree in Electrical and Computer Engineering with the Herbert Wertheim College of Engineering, University of Florida, Gainesville, FL, USA.
Prior to his graduate studies, he worked as an image algorithm engineer in the imaging industry, focusing on AI-ISP, HDR imaging, and auto white balance for embedded camera systems.
His research interests include low-level vision, computational photography, vision-language models, and edge intelligence.
\end{IEEEbiography}

\begin{IEEEbiography}[{\includegraphics[width=1in,height=1.25in,clip,keepaspectratio]{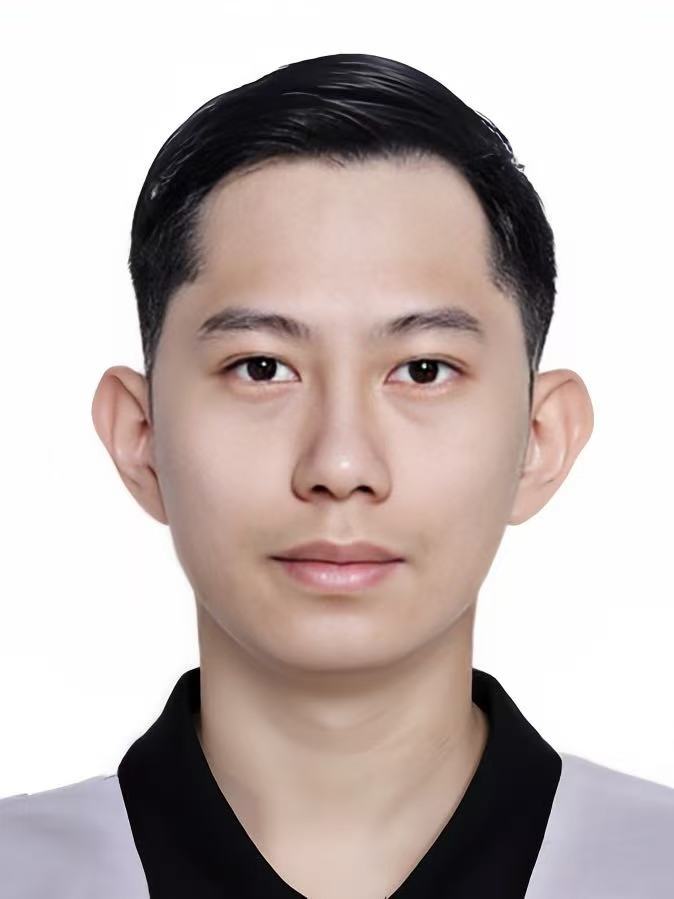}}]{Xinyang Wang}
Xinyang Wang received his B.E.
degree in 2025 from the Department of Electronic Engineering at Shantou University, China.
He is currently pursuing the M.E. degree in Electrical and Computer Engineering at the University of Florida.
His research interests include machine learning, computer vision, and embedded systems.
\end{IEEEbiography}

\begin{IEEEbiography}[{\includegraphics[width=1in,height=1.25in,clip,keepaspectratio]{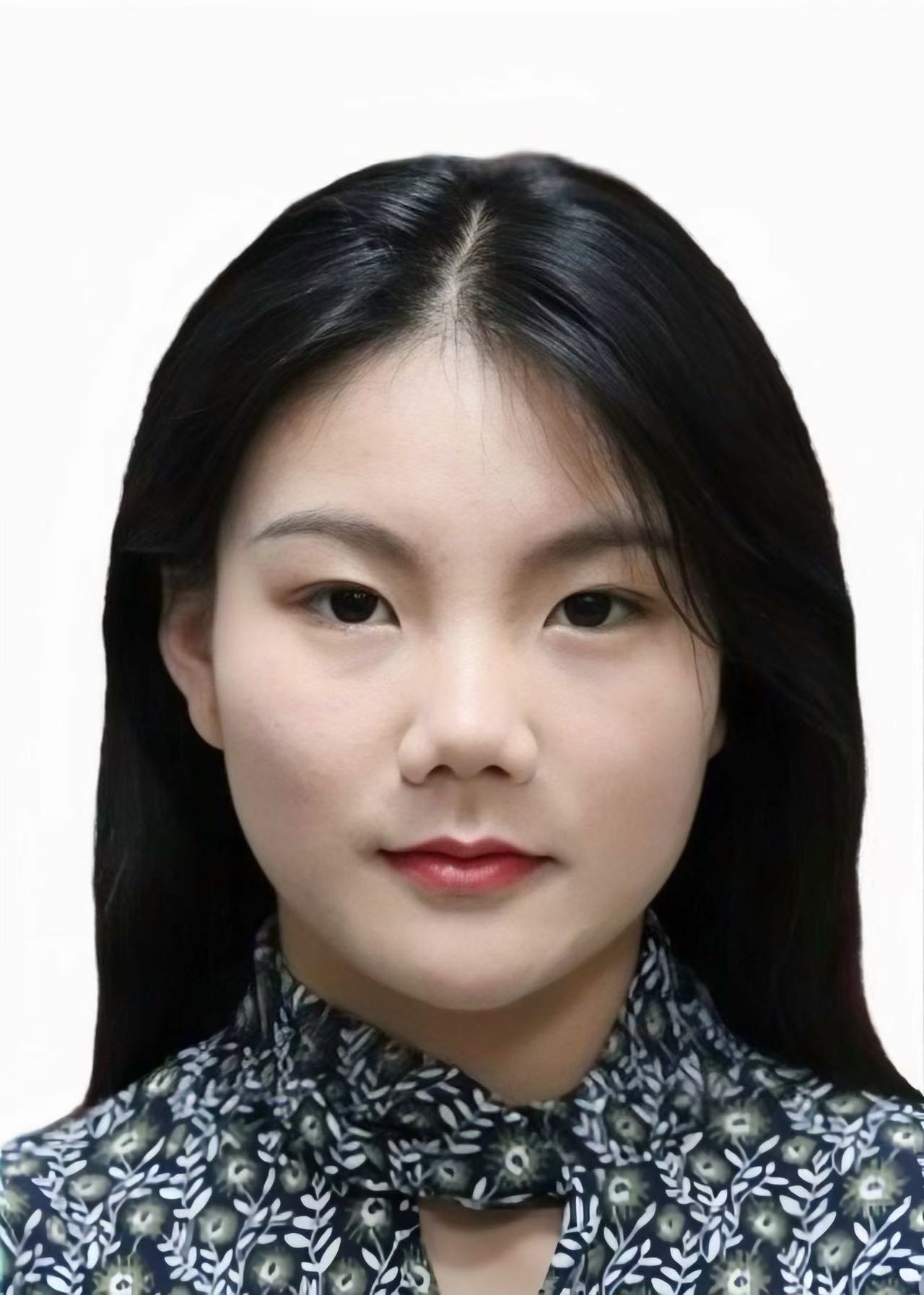}}]{Yahui Zhang}
Yahui Zhang is currently pursuing the Ph.D.
degree at the University of Southampton, U.K. Her research interests include AI-generated images and visual culture studies.
\end{IEEEbiography}

\begin{IEEEbiography}[{\includegraphics[width=1in,height=1.25in,clip,keepaspectratio]{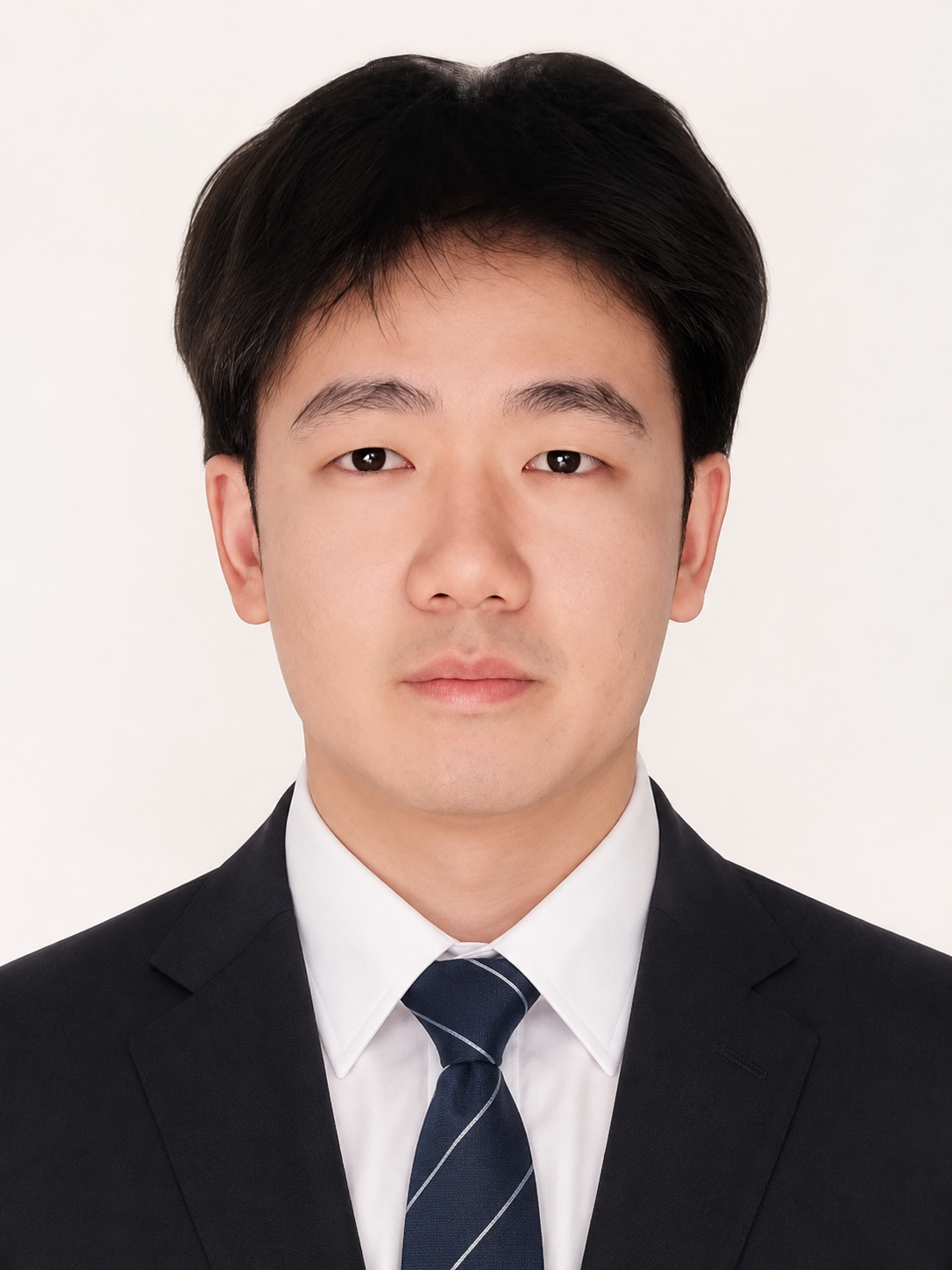}}]{Yuhan Chen}
Yuhan Chen received his master's degree in 2024 from the College of Mechanical Engineering at Chongqing University of Technology.
He is currently pursuing the Ph.D. degree in the College of Mechanical and Vehicle Engineering at Chongqing University, China.
His research interests include deep learning, low-level vision, and Gaussian splatting.
\end{IEEEbiography}

\begin{IEEEbiography}[{\includegraphics[width=1in,height=1.25in,clip,keepaspectratio]{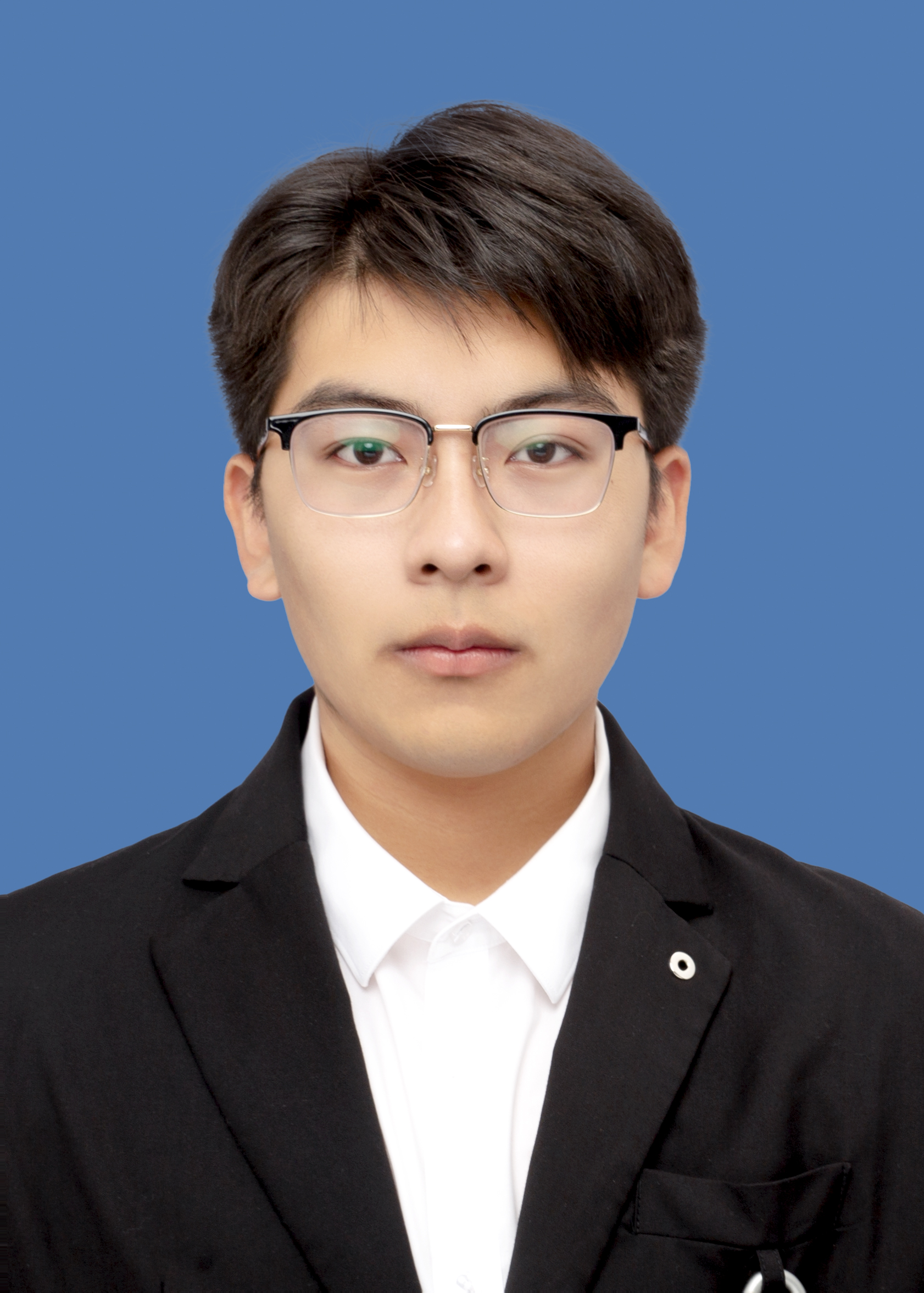}}]{Jiahong Guo}
Jiahong Guo is currently pursuing the B.S.
degree in Microbiology and Cell Science at the University of Florida, United States.
His research interests in artificial intelligence in medical imaging, computer vision, deep learning, and dental image analysis.
His current work focuses on intelligent medical imaging systems and the integration of artificial intelligence with dental and biomedical research.
\end{IEEEbiography}

\begin{IEEEbiography}[{\includegraphics[width=1in,height=1.25in,clip,keepaspectratio]{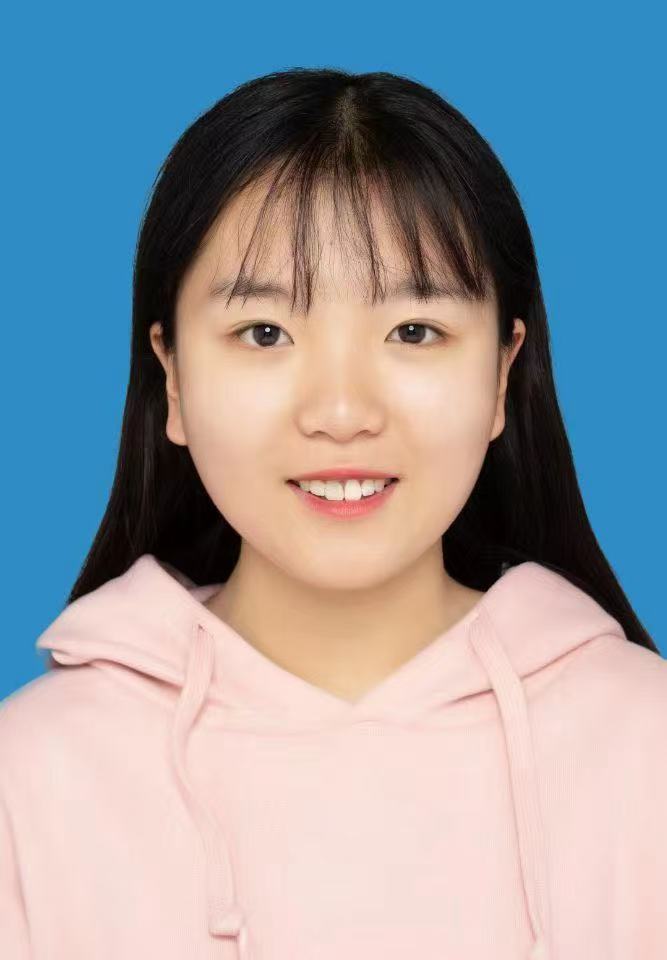}}]{Zhuohan Qin}
Zhuohan Qin is currently pursuing the M.S. degree in applied statistics with the School of Mathematics and Statistics, Qingdao University, Qingdao, China.
Her research interests include computer vision, video understanding, and procedural action assessment.
\end{IEEEbiography}

\begin{IEEEbiography}[{\includegraphics[width=1in,height=1.25in,clip,keepaspectratio]{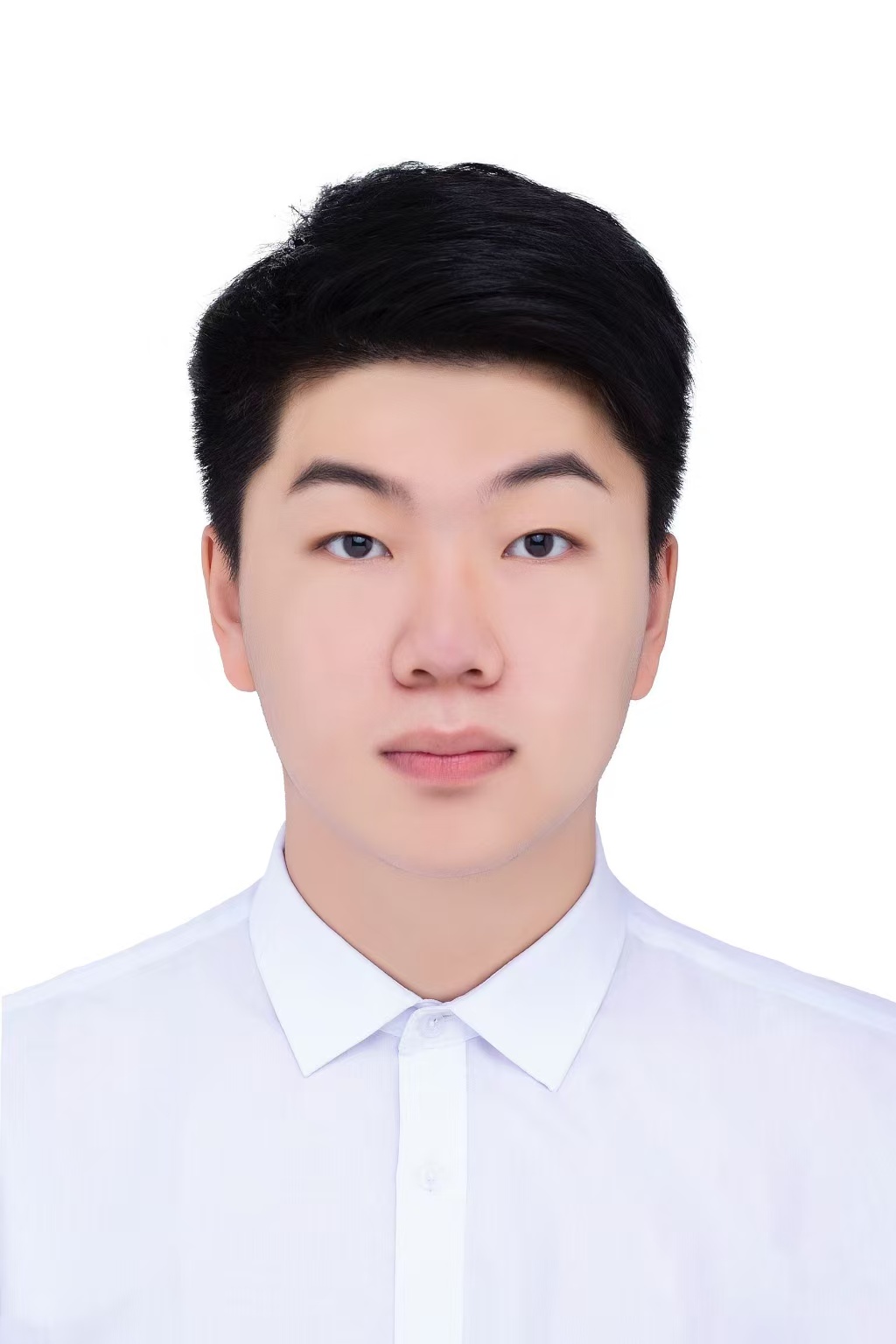}}]{Wenbo Ma}
Wenbo Ma is currently a Middleware Development Engineer with Intel Asia-Pacific Research \& Development Ltd., Shanghai, China, where he focuses on cross-ISA kernel development, low-level hardware optimization, and AI high-performance computing.
\end{IEEEbiography}
\vfill
\end{document}